%% file: main.tex
\PassOptionsToPackage{table,dvipsnames}{xcolor}
\documentclass[]{techreport}

\usepackage[utf8]{inputenc}
\usepackage[T1]{fontenc}
\usepackage{xspace}
\usepackage{graphicx}
\usepackage{amsmath}
\usepackage{amssymb}
\usepackage{booktabs}
\usepackage{url}
\usepackage{amsfonts}
\usepackage{nicefrac}
\usepackage{microtype}
\usepackage{xcolor}

\usepackage{eccvabbrv}

\input{preamble}

\title{\whatmovestitle{} Localized Motion Representations for Compositional Scene Control}

\author{
    {\large
    \textbf{Frank Fundel}\NoHyper\thanks{Equal contribution.}\endNoHyper,
    \hspace{0.3em} \textbf{Malek Ben Alaya}\footnotemark[1],
    \hspace{0.3em} \textbf{Thomas Ressler-Antal}\footnotemark[1], \\
    \textbf{Stefan Andreas Baumann},
    \hspace{0.3em} \textbf{Bj\"orn Ommer}}\\
    CompVis @ LMU Munich, Munich Center for Machine Learning (MCML)
}

\abstract{%
    \input{sec/0_abstract}

    \textbf{Project Page:} \url{https://compvis.github.io/WhatMoves}\\
    \textbf{Code:} \url{https://github.com/CompVis/WhatMoves}
    \vspace{-3.5em}
}

\begin{document}
\maketitle



\begin{figure}[H]
    \centering
    \includegraphics[width=\linewidth]{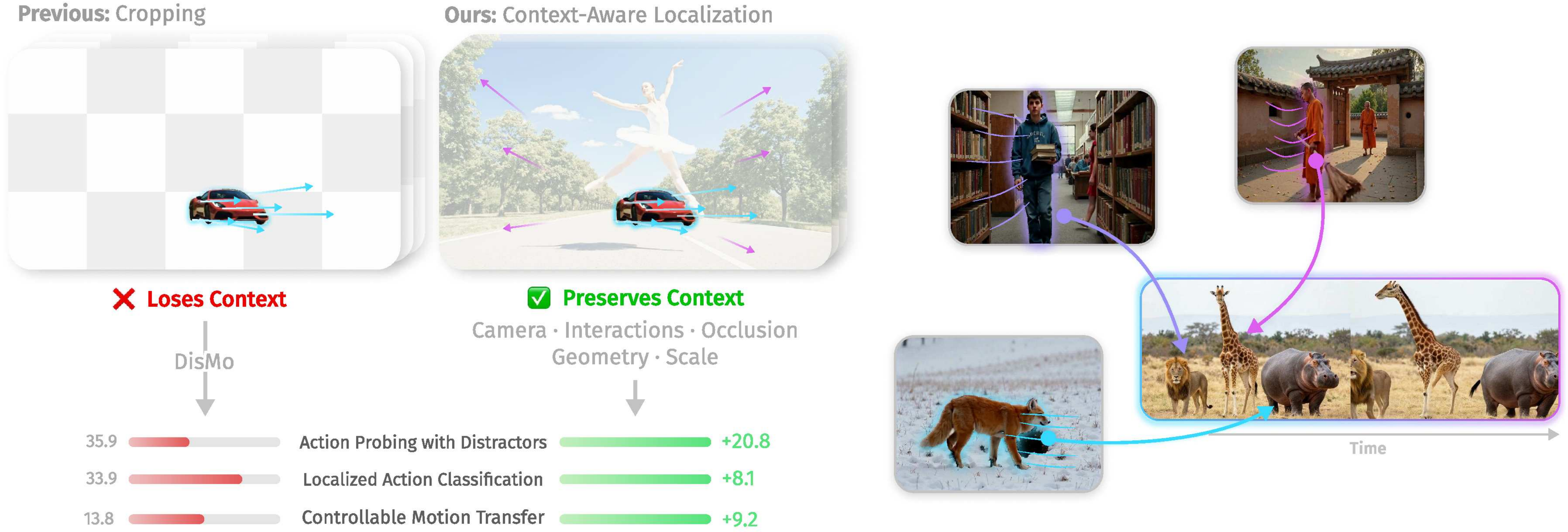}
    \vspace{0.5em}
    \caption{
        \textbf{Localized motion representations enable context-aware scene composition.}
        \textbf{(Left)} Our method extracts motion from selected regions while preserving global contextual cues such as camera motion, interactions, and scale, unlike cropping-based localization. \textbf{(Right)} These localized representations can be transferred from multiple source videos to target objects, enabling the composition of distinct motions within a single scene.
    }
    \label{fig:title}
\end{figure}
\setcounter{footnote}{0}

\input{sec/1_intro}
\input{sec/2_related}
\input{sec/3_method2}
\input{sec/4_experiments}

\input{sec/5_conclusion}

{
    \small
    \bibliographystyle{ieeenat_fullname}
    \bibliography{main}
}

\input{sec/X_supp}

\end{document}

%% file: preamble.tex
\usepackage{xparse}
\usepackage{adjustbox}
\usepackage{float}
\usepackage{wrapfig}
\usepackage{makecell}
\usepackage{enumitem}
\usepackage{comment}
\usepackage{rotating}
\usepackage{array}
\usepackage{tabularx}
\usepackage{siunitx}
\usepackage{bigdelim}
\usepackage{mathtools}
\usepackage{tikz}

\usetikzlibrary{arrows.meta,calc,positioning,fit,backgrounds}

\definecolor{wmgradcyan}{HTML}{37DAFF}
\definecolor{wmgradlavender}{HTML}{9E8EF5}
\definecolor{wmgradmagenta}{HTML}{DC5FEF}
\newcommand{\whatmovestitle}{%
  \mbox{\textit{%
    \textcolor{wmgradcyan}{W}%
    \textcolor{wmgradcyan!80!wmgradlavender}{h}%
    \textcolor{wmgradcyan!60!wmgradlavender}{a}%
    \textcolor{wmgradcyan!40!wmgradlavender}{t}%
    \textcolor{wmgradlavender}{\ }%
    \textcolor{wmgradlavender!80!wmgradmagenta}{M}%
    \textcolor{wmgradlavender!60!wmgradmagenta}{o}%
    \textcolor{wmgradlavender!40!wmgradmagenta}{v}%
    \textcolor{wmgradlavender!20!wmgradmagenta}{e}%
    \textcolor{wmgradmagenta}{s}%
    \textcolor{wmgradmagenta}{?}%
  }}%
}

\RenewDocumentCommand{\paragraph}{s m}{\vspace{.25em}\noindent\textbf{#2\IfBooleanF{#1}{.}}}

\newcommand{\tablescalebox}[1]{\adjustbox{max width=\linewidth}{\scalebox{.85}{#1}}}

\newcommand{\spantextfullarrow}[3]{%
  \multicolumn{\numexpr#2-#1+1\relax}{c}{\phantom{#3}\llap{\raisebox{-.3em}{#3}}} \\
  \cmidrule(lr){#1-#2}
  \ifnum#1>1
    \multicolumn{\numexpr#1-1\relax}{c}{}%
  \fi
  &\multicolumn{\numexpr#2-#1+1\relax}{r@{}}{%
    \llap{%
      \raisebox{1.2ex}[0pt][0pt]{%
        \tikz[baseline=-.65em]{\draw[->,line width=0.4pt](0,0)--(0.01,0);\node[] at (.07em,0) {};}%
      }%
      \hspace{\tabcolsep}%
    }%
  }\\[-2.35em]
}

\crefname{appsec}{Supp.\ Sec.}{Supp.\ Secs.}
\Crefname{appsec}{Supp.\ Section}{Supp.\ Sections}
\crefname{appfig}{Supp.\ Fig.}{Supp.\ Figs.}
\Crefname{appfig}{Supp.\ Figure}{Supp.\ Figures}
\crefname{apptab}{Supp.\ Tab.}{Supp.\ Tabs.}
\Crefname{apptab}{Supp.\ Table}{Supp.\ Tables}
\crefname{appeq}{Supp.\ Eq.}{Supp.\ Eqs.}
\Crefname{appeq}{Supp.\ Equation}{Supp.\ Equations}

%% file: sec/0_abstract.tex
Real-world dynamics are inherently compositional: multiple entities move simultaneously within a shared scene, each exhibiting distinct motion patterns. Yet current motion representation models entangle the dynamics of different entities, without explicitly capturing localized motion for each individually. Crucially, motion is defined relative to a global reference frame, including camera motion and scene layout. However, localized embeddings are often computed from cropped images or obtained by masking features after encoding, discarding the context needed to interpret motion. To address this, we introduce a promptable localized motion representation that produces persistent embeddings for user-specified regions defined by spatial masks. Rather than cropping the input or masking features, our model processes the full video and conditions motion encoding directly on the queried region. This yields temporally consistent, region-addressable embeddings that isolate local dynamics while retaining the global context required for disambiguation.
We demonstrate object-level motion transfer, enabling controlled composition of dynamic scenes.
Beyond generative control, our embeddings support localized action classification in multi-actor videos. Across both tasks, our approach improves controllability and outperforms global representations localized through cropping or post-hoc masking.

%% file: sec/1_intro.tex
\section{Introduction}
\label{sec:intro}
Motion is a defining characteristic of the visual world. Real-world scenes are inherently compositional, consisting of multiple entities whose dynamics unfold simultaneously within a shared environment. A driving scene contains vehicles, pedestrians, and background motion with distinct temporal patterns; human-object interactions involve coordinated, yet separable movements of body parts and manipulated objects. Modeling such dynamics requires representations that are localized, persistent, and semantically grounded. Despite rapid advances in video generation and motion control, most existing approaches represent motion either globally or at the pixel level. Global motion embeddings entangle the dynamics of all entities into a single latent code~\cite{ressler-antal2025dismo}, while dense optical flow or trajectory fields encode motion locally but lack semantic structure and compositional interpretability. As a result, motion remains difficult to isolate, recombine, or selectively manipulate at the object level. This limitation becomes particularly apparent in tasks such as motion transfer and video editing. Existing methods often extract motion globally and reapply it to new scenes~\cite{ressler-antal2025dismo,gu2025diffusion,shi2024motion,DeT,ma2026fastvmt,sheng2026flexam}, leading to motion leakage, spatial ambiguity, or incorrect attachment to unintended regions. While these approaches demonstrate impressive generative quality, they lack an explicit, semantic representation of who is moving and how. A seemingly straightforward solution is to localize motion by applying global motion encoders to spatial crops of a video. Yet motion is inherently relational: its meaning depends on global reference frames and interactions between entities. When this context is removed, the motion signal within a crop may no longer uniquely determine the underlying dynamics. Consequently, embeddings extracted from isolated regions are not guaranteed to represent well-defined object-level motion.

Our approach operates directly on the full input video while leveraging spatiotemporal spatial masks to localize motion signals. Instead of cropping regions, we retain the entire scene context and apply region-specific masks over time within the encoder. A motion encoder processes the masked video features to produce compact entity-level motion embeddings, while a motion decoder reconstructs the corresponding localized point trajectories, encouraging spatial specificity and disentanglement from static appearance. By preserving global context during encoding and enforcing localization through masking, our method yields motion representations that are both contextually grounded and spatially isolated. This structured representation naturally enables compositional video generation and control. Motion from individual entities can be selectively extracted, recombined, and reattached to arbitrary targets to animate a scene. Beyond motion transfer, the learned embeddings support localized action classification, highlighting their role as general representations for motion understanding and generation.

Through extensive experiments, we demonstrate that our approach enables precise spatial controllability and structured motion composition in generative settings. In localized motion transfer experiments, our method supports accurate extraction and reattachment of region-specific dynamics while significantly reducing motion leakage compared to global motion conditioning. It further enables compositional recombination of motions across scenes and more consistent cross-view transfer than track-conditioned models that rely on viewpoint-specific trajectories. Beyond generation, we evaluate the learned motion embeddings on motion understanding tasks. In localized action classification, our representations capture spatially specific motion patterns that generalize across actors and scenes, outperforming global video embeddings localized via cropping or post-hoc masking. Together, these results highlight the importance of encoding local motion while preserving full-scene context for both controllable generation and motion understanding.

\noindent Our main contributions are summarized as follows:

\begin{itemize}[leftmargin=15pt]
\item We introduce a promptable localized motion representation that encodes persistent, region-specific motion embeddings conditioned on spatial regions.

\item We show that encoding local motion requires full-scene context, and demonstrate through localized action classification that crop-based or post-hoc masking approaches are insufficient.

\item We demonstrate that the learned motion embeddings enable object-level motion transfer as well as selective extraction and recombination of region-level dynamics, supporting controllable and compositional scene animation in image-to-video settings.

\end{itemize}

%% file: sec/2_related.tex
\section{Related Work}
\label{sec:related}
\paragraph{Video Generation Models}
Diffusion-based generative models have driven rapid progress in video synthesis. Building on advances in text-to-image generation~\cite{ho2020denoising,song2020score,rombach2022ldm}, many works extend pretrained image backbones with temporal modules to generate coherent video sequences~\cite{Chen_2024_CVPR,guo2024animatediffanimatepersonalizedtexttoimage}. Scaling model capacity and training data has further improved temporal consistency and semantic richness in text-to-video (T2V) and image-to-video (I2V) systems~\cite{brooks2024sora,blattmann2023stable,yang2024cogvideox,HaCohen2024LTXVideo,kong2024hunyuanvideo,wan2025wanopenadvancedlargescale}.
Despite these advances, most video generation systems control motion primarily through text prompts, which provide only coarse and indirect guidance over how objects move. We introduce an object-centric representation of motion that enables targeted manipulation of dynamics at the level of individual entities. Our approach can be integrated with existing video generation models, enabling more precise and structured control over motion without redesigning the underlying generative architecture.

\paragraph{Motion Control in Video Generation}
Recent work improves controllability in video diffusion models by introducing explicit motion guidance signals. A common approach relies on image-plane geometric control, where motion is specified through trajectories, bounding boxes, or structural signals such as optical flow and depth~\cite{wang2024motionctrl,wu2024draganything,yin2023dragnuwa,jain2024peekaboo,wang2024boximator,zhang2023controlvideo,esser2023structure}. Subsequent works extend this paradigm with trajectory-aware conditioning mechanisms and motion-guided attention modules that enable more precise point-level motion control and multi-entity interactions~\cite{geng2024motionprompting,wang2025ati,shin2025motionstreamrealtimevideogeneration,xiao2024trajectory,zhang2025tora,zhang2025tora2}. Other approaches guide generation directly using motion signals such as optical flow, temporally warped noise, or related geometric cues~\cite{koroglu2025onlyflow,burgert2025gowiththeflowmotioncontrollablevideodiffusion,zhou2025trackgoflexibleefficientmethod}.
While these approaches provide strong geometric control when the guidance signal aligns with the target layout, motion is typically represented in pixel space through trajectories, flow fields, or related spatial signals. As a result, the encoded dynamics remain tightly coupled to viewpoint, object layout, and scene configuration, making it difficult to transfer the same motion across changes in camera perspective, object shape, or scene structure.

A complementary line of work studies motion representation learning and motion transfer, where dynamics are extracted from a source video and applied to a different scene. Several diffusion-based text-to-video approaches leverage internal features of pretrained video diffusion models to encode and propagate motion across scenes~\cite{jeong2024vmc,yatim2024space,xiao2024video,ling2024motionclone,zhao2024motiondirector,pondaven2025videomotiontransferdiffusion,DeT}. More recent work explores motion-guided editing paradigms that modify subject appearance while preserving source motion~\cite{liu2025multimotionmultisubjectvideo,liu2025motionshot}. Other approaches focus on efficiency or representation abstraction, such as training-free motion transfer formulations or depth-aware motion embeddings~\cite{ma2026fastvmt,sheng2026flexam}. Methods operating on lower-level features often inherit geometric constraints from the source video, while higher-level feature-based approaches provide greater flexibility but may entangle motion with appearance, leading to content leakage. Recent work proposes more abstract motion embeddings to mitigate this entanglement~\cite{ressler-antal2025dismo,huberman2026semanticmomentstrainingfreemotionsimilarity}. However, these approaches typically represent motion globally or as latent feature dynamics within a generative backbone, without explicit object-level structure or spatial localization. Consequently, they offer limited control over where motion is extracted from and how it is distributed across entities in multi-subject scenes.

Our work differs in that we learn promptable, localized motion representations conditioned on spatial regions. Unlike trajectory-based control methods that rely on image-plane guidance signals, and motion transfer approaches that represent dynamics globally within a generative backbone, our approach models motion at the object level while preserving full-scene context.

%% file: sec/3_method2.tex
\section{Method}
\label{sec:method}

\begin{figure*}[t]
    \centering
    \includegraphics[width=0.9\linewidth]{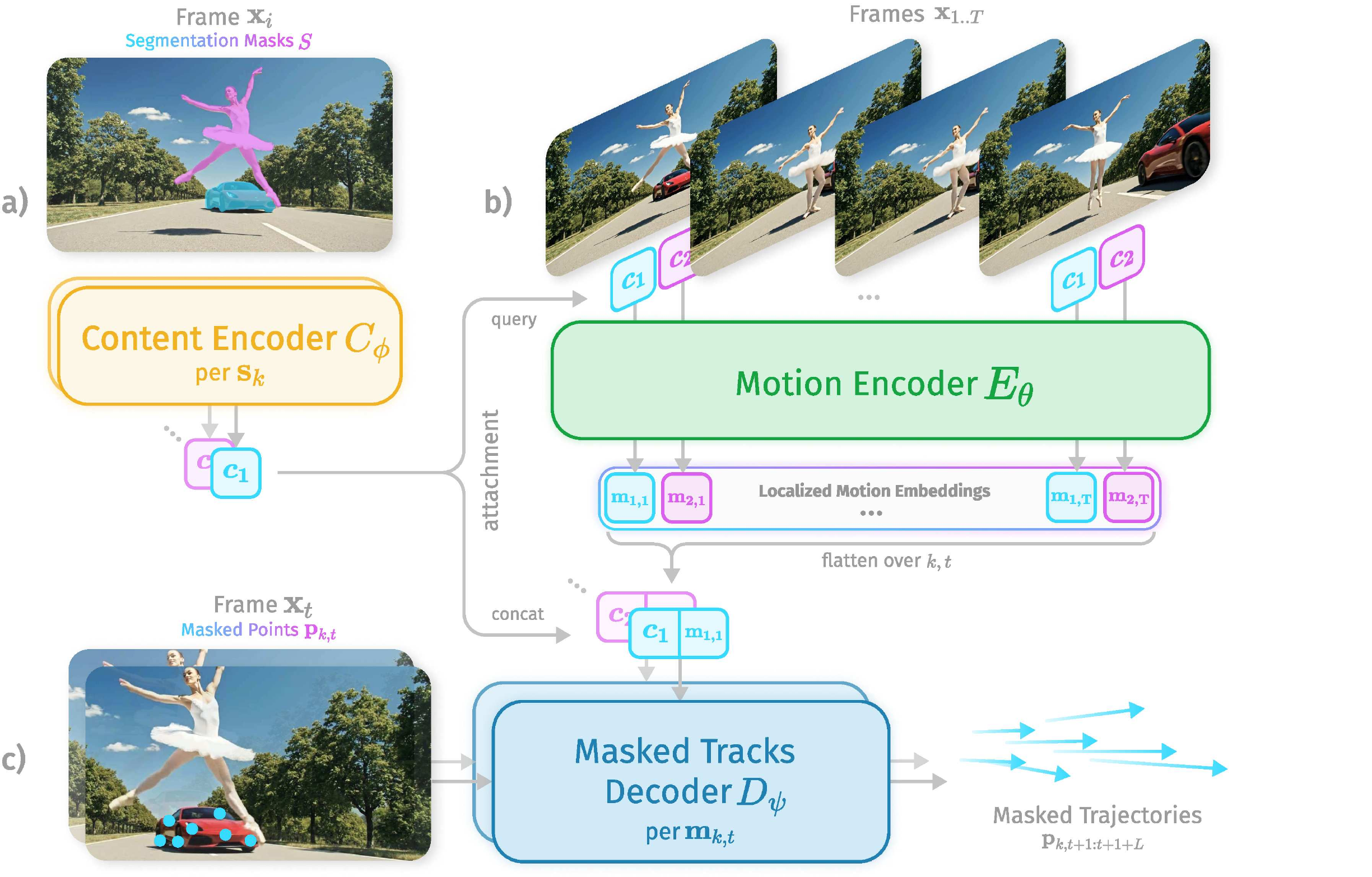}
    \caption{
    \textbf{Method Overview.}
        Given a video frame with corresponding object masks, the content encoder produces static region tokens for appearance grounding \textbf{(a)}. The motion encoder \textbf{(b)} extracts persistent, region-specific motion tokens while preserving global context. The masked flow decoder \textbf{(c)} predicts region-level point trajectories, enforcing localized motion supervision within masked areas.
    }
    \label{fig:method}
\end{figure*}

    Modeling video dynamics in a structured and controllable manner requires motion representations that are both spatially localized and grounded in global scene context. While global motion embeddings capture overall dynamics, they do not provide entity-level motion units that can be independently extracted and recombined. In this section, we first examine the challenges of localizing motion in video and motivate a context-aware formulation. We then introduce our promptable localized motion representation.

\subsection{Motivation: Localized Motion Requires Global Context}
A natural strategy for isolating entity motion is to encode motion from spatial crops. However, unlike appearance, motion is inherently \emph{relational}: its interpretation depends on global context, camera motion, and interactions between objects. For instance, an object moving away from the camera may produce the same local motion pattern as a camera zoom-out. Similarly, object motion and camera motion often become indistinguishable when observed in isolation. These ambiguities arise because spatial cropping removes the reference frame needed to interpret motion. Consequently, crop-based motion encoders frequently learn ambiguous dynamics or collapse to dominant camera motion. Reliable entity-level motion representations therefore require localization \emph{without discarding global context}. This observation motivates our formulation: the encoder processes the \emph{entire video}, while entity-specific motion is extracted through structured prompting and masked supervision.

\subsection{Promptable Localized Motion Embeddings}
Building on this observation, we introduce a promptable localized motion representation that processes the full video while extracting region-specific motion embeddings conditioned on spatiotemporal masks. Each embedding captures the temporal dynamics of a queried entity and remains persistent over time. By preserving global context during encoding and enforcing region-level aggregation through masking, our framework produces motion representations that are both spatially precise and contextually grounded.\\
Consider a video $V = \{\mathbf{x}_t\}_{t=1}^{T}$ with $T$ frames. In the first frame $\mathbf{x}_1$, we are given region masks $S := \{\mathbf{s}_k\}_{k=1}^{K}$. Our goal is to extract persistent, region-specific motion tokens $\mathbf{m}_k := \{\mathbf{m}_{k,t}\}_{t=1}^{T}$, where $\mathbf{m}_{k,t}$ describes the motion of region $k$ at time $t$. We tackle this problem using three components:
(1) a \textbf{motion encoder} that extracts localized motion tokens from the full video,
(2) a \textbf{content encoder} that provides appearance grounding for the queried entity, and
(3) a \textbf{masked tracks decoder} that supervises motion tokens through future trajectory prediction. An overview of our framework is depicted in \cref{fig:method}.

\paragraph{Motion Encoder}
The motion encoder $E_\theta$ maps the full video $V$ and a set of entity content embeddings
$\mathbf{c}=\{\mathbf{c}_k\}_{k=1}^{K}$ to temporally evolving motion tokens
$\mathbf{m}=\{\mathbf{m}_{k}\}_{k=1}^{K}$, i.e.,
$E_\theta:(V,\mathbf{c})\mapsto\mathbf{m}$.
Importantly, $E_\theta$ observes the entire scene, allowing each entity-specific motion token to use global context such as camera motion, occlusions, and object interactions. The content embeddings serve as region-specific queries, encouraging $\mathbf{m}_{k,t}$ to capture the motion of the corresponding entity while reducing leakage from other regions.

\paragraph{Content Encoder} The content encoder $C_\phi$ provides appearance grounding by mapping a frame $\mathbf{x}_i$ and region mask $\mathbf{s}_k$ to a content token $\mathbf{c}_k$, \textit{i.e.}, $C_\phi : (\mathbf{x}_i, \mathbf{s}_k) \rightarrow \mathbf{c}_k$. This token captures the semantic identity of the prompted region and is time-invariant. It queries the motion encoder to specify the entity whose motion should be extracted and is passed to the tracks decoder to reattach motion embeddings to the correct region, discouraging them from encoding spatial identity.

\paragraph{Masked Tracks Decoder and Objective}
We supervise the motion tokens with a masked tracks decoder $D_\psi$. For each entity $k$, sparse query points are sampled only inside its region $S_k$ and tracked through the video, yielding trajectories $\mathbf{p}_{k,1:T}$. At a valid timestep $t$, the decoder receives the current frame $\mathbf{x}_t$, query positions $\mathbf{p}_{k,t}$, the content embedding $\mathbf{c}_k$, and the motion token $\mathbf{m}_{k,t}$, and predicts future track positions $\mathbf{p}_{k,t+1:t+L}$.
Although this conditional trajectory distribution could be modeled generatively, e.g. with a flow matching decoder, we use a deterministic decoder trained with an $\ell_2$ reconstruction loss:
\[
\mathcal{L}(\theta,\psi)
=
\mathbb{E}_{V,k,t}
\left[
\left\|
D_\psi(\mathbf{x}_t,\mathbf{p}_{k,t},\mathbf{c}_k,\mathbf{m}_{k,t})
-
\mathbf{p}_{k,t+1:t+L}
\right\|_2^2
\right],
\]
evaluated only for valid future track positions within a region, encouraging each motion token to encode localized motion rather than unrelated scene dynamics.

\subsection{Localized Motion-Guided Video Generation}
\label{subsec:video_model_conditining}
\begin{figure*}[t]
    \centering
    \includegraphics[width=0.9\linewidth]{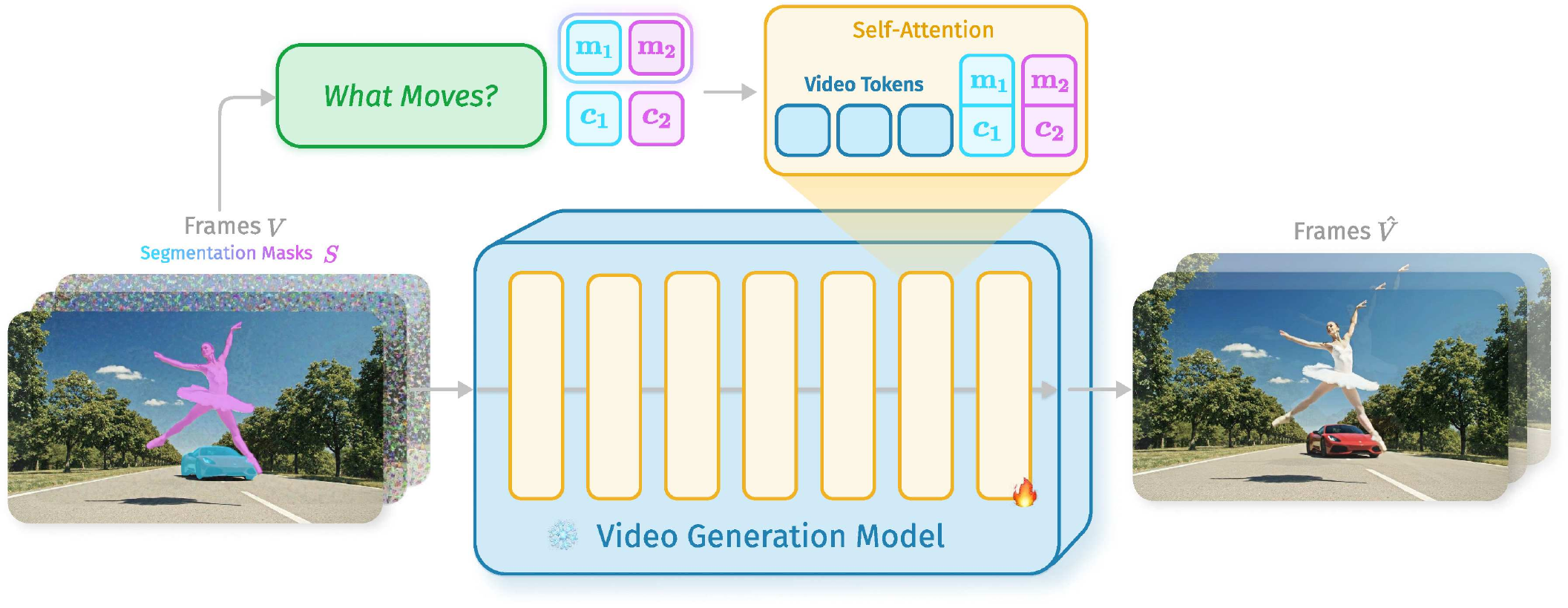}
    \caption{\textbf{Video Model Conditioning.} With our localized representations, we fine-tune a large video generation model. For this, we append to each temporal self-attention layer of the the model, the respective motion embeddings $\mathbf{m}_{1:K,t}$ and content embeddings $\mathbf{c}_{1:K}$ for all regions $K$.
}
    \label{fig:method2}
\end{figure*}

Once trained, the motion encoder extracts region-specific motion tokens from a source video by querying a spatial region. These tokens capture the dynamics of the selected entity over time and enable localized motion transfer, where motion from a source region is attached to a specified object in a target scene. To generate videos following these dynamics, we condition a pretrained video diffusion model on sequences of localized motion tokens. For each region $k$ and timestep $t$, we concatenate the motion tokens $m_k,t$ with their corresponding content token $c_k$,
\begin{equation}
    \mathbf{q}_{k,t} = [\mathbf{m}_{k,t}; \mathbf{c}_k],
\end{equation}
and append the resulting conditioning tokens to the video token sequence $\mathbf{V}_t = \{\mathbf{v}_{t,n}\}^N_{n=1}$ processed by the model's self-attention layers:
\begin{equation}
    \mathbf{V}'_t
    =
    [\mathbf{V}_t, \mathbf{q}_{1..K,t}].
\end{equation}
The backbone is fine-tuned using LoRA~\cite{hu2021loralowrankadaptationlarge} to incorporate the additional conditioning while preserving the pretrained weights. This fine-tuning approach is illustrated in \cref{fig:method2}.

%% file: sec/4_experiments.tex
\section{Experiments}
\label{sec:experiments}
To evaluate the properties of our motion representations, we conduct experiments across several settings. We study controllable motion transfer to assess whether motion can be transferred across entities and scenes enabling compositional scene control, localized action classification to test whether motion can be extracted for a queried subject in the presence of distractors, and global motion understanding tasks to evaluate whether the representations retain semantic motion properties beyond localized queries.

\paragraph{Implementation Details}
We implement our motion encoder as a 3D Vision Transformer~\citep[][$86$M parameters]{dosovitskiy2021vit}, preceded by a DINOv2-B~\citep{oquab_dinov2_2023} frame embedding stage. The masked flow decoder is also implemented as a Transformer[$86$M parameters]. We jointly train the encoder and decoder using AdamW~\citep{loshchilov2019decoupledweightdecayregularization} for $600$k steps with a batch size of $128$ on open-world videos from OpenVid-1M~\citep{nan2025openvidm} and an internally collected dataset, enabling the model to learn general-purpose motion representations. For motion transfer tasks, we utilize CogVideoX-5B~\citep{yang2025cogvideoxtexttovideodiffusionmodels} as a backbone. 
We provide further implementation details in the supplementary material.

\subsection{Motion Transfer}
\paragraph{Localized Motion Attachment}
In this experiment, we evaluate whether our framework can attach motion from a source region to a specified target region. To this end, we construct a benchmark that tests whether motion transfer methods can control \emph{which} subject receives transferred motion. Each source video mostly contains a single subject, while the target initial frame always contains multiple subjects. Our method transfers motion from the source subject to a designated target subject, whereas baseline methods without subject-level prompting perform global motion transfer without explicit target selection. This setting enables direct evaluation of whether motion can be accurately directed to a specified subject while suppressing unintended propagation to others. We select source videos from the MTBench dataset~\cite{DeT}, originally gathered from DAVIS~\cite{ponttuset20182017davischallengevideo} and VOS~\cite{xu2018youtubevoslargescalevideoobject}, and extract subject masks using the promptable segmentation model SAM2~\cite{ravi2024sam2segmentimages}. For each source video, we generate multiple target initial frames using Z-Image Turbo~\cite{imageteam2025zimageefficientimagegeneration}.\\

We quantitatively compare methods using two established global metrics: \emph{Temporal Consistency}, measured via alignment of CLIP embeddings between consecutive generated frames; and \emph{Driving Video Similarity}, computed as the alignment between CLIP embeddings of source and generated video frames. To specifically assess subject-level controllability, we compute an \emph{in-region fidelity} score using a tracking-based Chamfer distance between the source subject trajectories and the trajectories of the designated target subject in the generated video. Higher values indicate more accurate transfer of motion to the intended subject. To quantify unintended motion propagation, we additionally compute an \emph{out-of-region leakage} score using the same Chamfer-based metric between the source subject trajectories and all non-target subject trajectories in the generated video. Lower values indicate better suppression of motion leakage.\\

\begin{figure*}[t]
    \centering
    \includegraphics[width=0.9\linewidth]{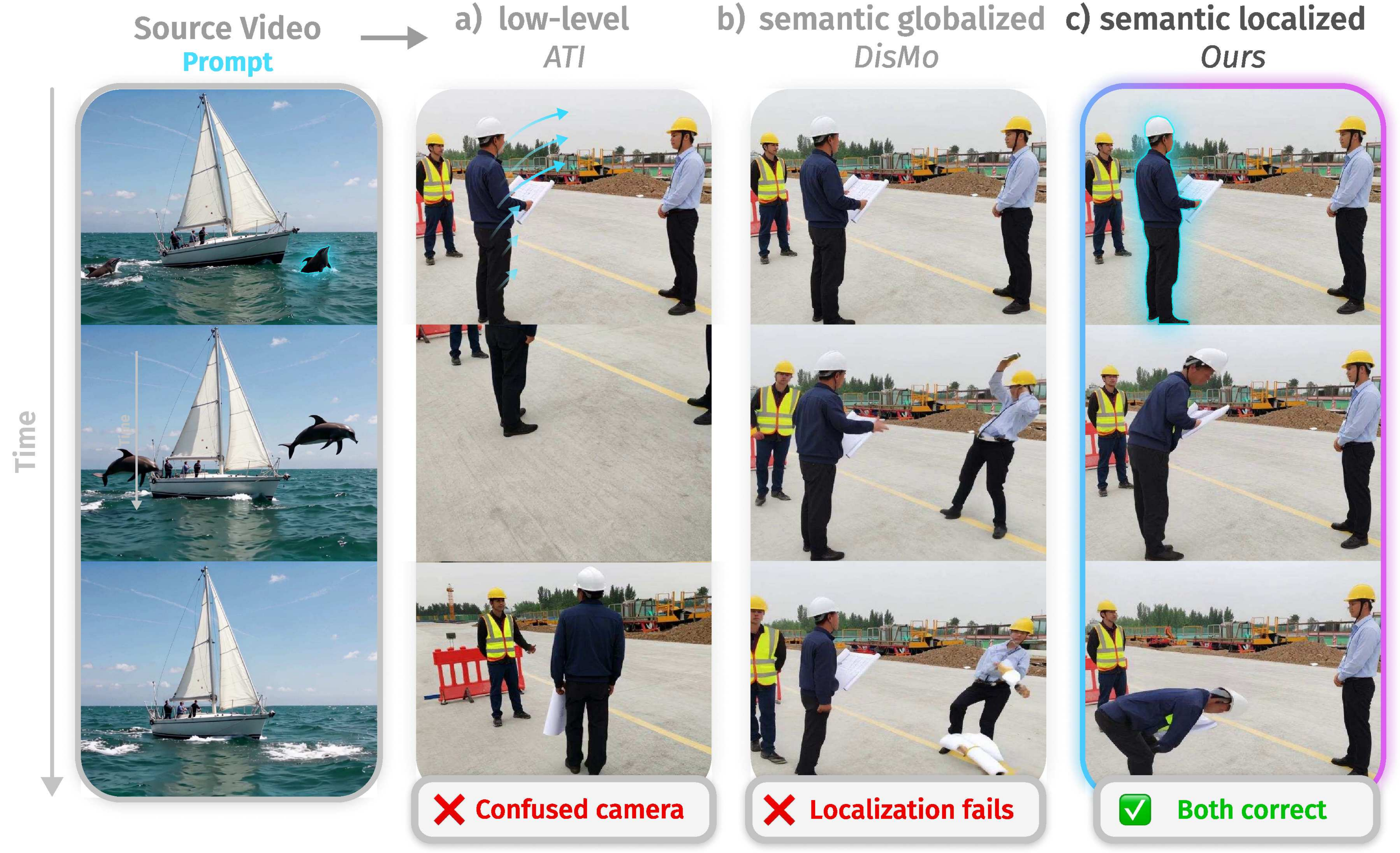}
    \caption{\textbf{Motion Transfer Paradigms.} Existing methods are typically either localized or semantic. \textbf{(a)} Low-level approaches transfer exact motion and implicitly assume identical structure; while localized, they do not generalize across different structures. \textbf{(b)} Semantic motion transfer methods generalize motion across structures but operate globally and lack localization. \textbf{(c)} Our approach enables motion transfer that is both semantic and localized, bridging the two paradigms.}
    \label{fig:comp}
\end{figure*}

We compare against two classes of motion transfer methods: low-level approaches, which provide localized control but rely on low-level motion signals such as tracks, and semantic motion transfer methods, which utilize higher-level motion representations but operate globally.
Naturally, low-level methods often achieve strong scores on motion fidelity metrics such as in-region motion fidelity, since they directly follow the source trajectories. However, these approaches do not generalize well across viewpoint changes or structural differences between source and target subjects.
In contrast, semantic methods capture higher-level motion patterns but typically lack spatial control. This leads to higher motion leakage, as motion is transferred to arbitrary subjects. We illustrate these trade-offs qualitatively in \cref{fig:comp}.
Our approach combines both properties by enabling localized attachment of semantic motion. As shown in \cref{tab:motion_metrics}, our method achieves strong in-region motion fidelity while substantially reducing motion leakage compared to global semantic conditioning. Although ATI \cite{wang2025ati} achieves the highest fidelity, it exhibits significantly higher out-of-region leakage.
Compared to the semantic baseline DisMo \cite{ressler-antal2025dismo}, our method improves in-region motion fidelity while improving localization. Overall, our approach achieves a favorable trade-off between motion fidelity and spatial isolation. Since our representation captures semantic motion rather than raw trajectories while remaining spatially localized, it generalizes to transfers across viewpoint changes and structural differences between source and target subjects.

\begin{table*}[t]
    \centering
    \caption{\textbf{Controllable Motion Transfer.}
    We report temporal consistency, source similarity, in-region motion fidelity, 
    out-of-region motion leakage, and motion selectivity, defined as the difference 
    between fidelity and leakage. Our method outperforms semantic motion transfer 
    approaches and achieves the best overall motion selectivity, demonstrating the 
    strongest balance between accurately transferring motion to the target region 
    and preventing leakage to the rest of the scene.
    }
    \label{tab:motion_metrics}
    \tablescalebox{
    \begin{tabular}{l@{\hskip 1.6em}l>{\color{gray}}c>{\color{gray}}c ccc}
    & & \multicolumn{2}{c}{\color{gray} video model properties}
    & \multicolumn{3}{c}{motion transfer properties} \\
    \cmidrule(lr){3-4} \cmidrule(lr){5-7}
    \toprule
    & \textbf{Model} &
    \makecell{\textbf{Temporal}\\\textbf{Consistency}}$\uparrow$ &
    \makecell{\textbf{Source}\\\textbf{Similarity}}$\uparrow$ &
    \makecell{\textbf{In-Region}\\\textbf{Motion Fidelity}}$\uparrow$ &
    \makecell{\textbf{Out-of-Region}\\\textbf{Motion Leakage}}$\downarrow$ &
    \makecell{\textbf{Motion}\\\textbf{Selectivity}}$\uparrow$ \\
    \midrule
    \multirow{3}{1.2em}{\rotatebox{90}{\scriptsize low-level}}
    & ATI \cite{wang2025ati} (Wan 2.1-14B \cite{wan2025wanopenadvancedlargescale})
    & 0.9725 & 0.5930 & \textbf{0.8111} & 0.6108 & 0.2003 \\
    & WanMove \cite{chu2025wanmovemotioncontrollablevideogeneration} (Wan 2.1-14B)
    & \underline{0.9738} & 0.5949 & \underline{0.7438} & \underline{0.5608} & 0.1830 \\
    & Tora \cite{zhang2025tora} (CogVideoX-5B \cite{yang2025cogvideoxtexttovideodiffusionmodels})
    & \textbf{0.9758} & 0.5807 & 0.7021 & \textbf{0.4738} & \underline{0.2283} \\
    \midrule
    \multirow{2}{1.2em}{\rotatebox{90}{\scriptsize\strut \shortstack{sem-\\antic}{\hskip .15em}}}
    & DisMo \cite{ressler-antal2025dismo} (CogVideoX-5B)
    & \underline{0.9657} & \textbf{0.5947} & \underline{0.6981} & \underline{0.5604} & 0.1377 \\
    & \textbf{Ours} (CogVideoX-5B)
    & \textbf{0.9690} & \underline{0.5888} & \textbf{0.7462} & \textbf{0.5158} & \textbf{0.2304} \\
    \bottomrule
    \end{tabular}
    }
\end{table*}

\paragraph{Composing Scenes}
\begin{figure*}[t]
    \centering
    \includegraphics[width=\linewidth]{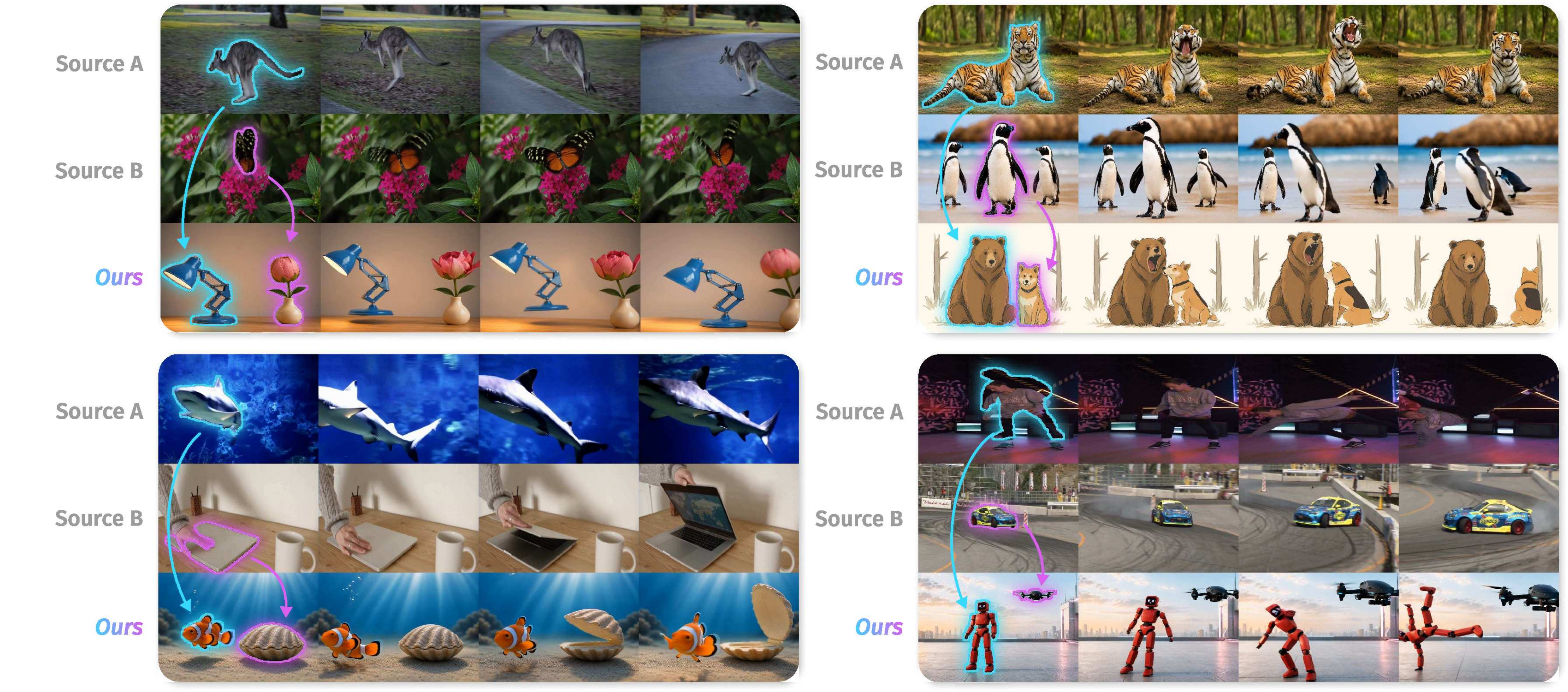}
    \caption{\textbf{Composite Qualitative Examples.} We provide several examples, showcasing our capability to compose scenes from localized motion embeddings of multiple source videos. Our model correctly isolates and attaches the motion of multiple subjects into one coherent scene.}
    \label{fig:qual}
    \vspace{-1em}
\end{figure*}
Beyond transferring motion to a single subject, compositional scene generation requires the ability to combine motion from multiple sources and assign them to different subjects within the same scene. To evaluate this capability, we extend our evaluation to a \emph{multi-source} setting where motion is extracted from multiple source videos and attached to different target subjects in a shared target frame. Each source video provides motion for a distinct subject, and our method is prompted to compose these motions within a single generated scene. \\

As shown in \cref{tab:multi_source_motion_metrics}, our method shows a competitive in-region motion fidelity, while substantially reducing out-of-region motion leakage compared to all baselines. This indicates that our localized motion representations enable controlled composition of multiple motion sources within the same scene, allowing motion to be selectively attached to the intended subjects without affecting others. Qualitative examples are shown in the supplementary material. \\

We additionally conduct a blinded user study using non-cherry-picked examples, with the full experimental setup provided in the supplementary material. Participants independently assess each generated video along three dimensions: overall visual realism, the quality and spatial locality of the transferred motion, and the presence of visible artifacts. \cref{tab:human_study} shows that across the evaluated methods, our approach is consistently preferred for both realism and motion transfer quality. In particular, it outperforms the next-best baseline by at least $27.78\%$ in realism and $9.8\%$ in motion transfer quality, indicating that the generated videos are perceived as both more visually plausible and more effective at transferring motion to the intended target region.

\begin{table}[t]
\centering
\caption{\textbf{Human preference study.}
Percentage of votes for realism, motion transfer quality and locality, and the absence of visible artifacts.}
\label{tab:human_study}
\adjustbox{max width=\linewidth}{
\begin{tabular}{lccccc}
\toprule
\textbf{Model}
& \textbf{Backbone}
& \textbf{\#Params}
& \textbf{Realism $\uparrow$}
& \textbf{Transfer $\uparrow$}
& \textbf{Fewer Artifacts $\uparrow$} \\
\midrule
ATI \cite{wang2025ati}
& WAN2.1 \cite{wan2025wanopenadvancedlargescale}
& 14B
& 1.96
& 4.90
& 5.23 \\
WanMove \cite{chu2025wanmovemotioncontrollablevideogeneration}
& WAN2.1
& 14B
& 14.05
& 14.38
& 14.38 \\
Tora \cite{zhang2025tora}
& CogVideoX \cite{yang2025cogvideoxtexttovideodiffusionmodels}
& 5B
& 13.73
& 17.32
& 15.69 \\
DisMo \cite{ressler-antal2025dismo}
& CogVideoX
& 5B
& 21.24
& 26.80
& 20.26 \\
\textbf{Ours}
& CogVideoX
& 5B
& \textbf{49.02}
& \textbf{36.60}
& \textbf{44.44} \\
\bottomrule
\end{tabular}}
\end{table}

\begin{table*}[t]
\centering
\caption{\textbf{Composable Motion Transfer.}
We report temporal consistency, in-region motion fidelity, out-of-region motion leakage, and motion selectivity when transferring motion from multiple source objects. Our approach substantially reduces motion leakage and achieves the best overall motion selectivity while remaining competitive in in-region motion fidelity.}
\label{tab:multi_source_motion_metrics}
\tablescalebox{
\begin{tabular}{l@{\hskip 1.6em}>{\color{gray}}c ccc}
\toprule
\textbf{Model} &
\makecell{\textbf{Temporal}\\\textbf{Consistency}}$\uparrow$ &
\makecell{\textbf{In-Region}\\\textbf{Motion Fidelity}}$\uparrow$ &
\makecell{\textbf{Out-of-Region}\\\textbf{Motion Leakage}}$\downarrow$ &
\makecell{\textbf{Motion}\\\textbf{Selectivity}}$\uparrow$ \\
\midrule
ATI \cite{wang2025ati} (Wan 2.1-14B \cite{wan2025wanopenadvancedlargescale})
& 0.9634 & \textbf{0.7148} & \underline{0.3857} & \underline{0.3291} \\

WanMove \cite{chu2025wanmovemotioncontrollablevideogeneration} (Wan 2.1-14B)
& \underline{0.9792} & \underline{0.7059} & 0.4110 & 0.2949 \\

Tora \cite{zhang2025tora} (CogVideoX-5B \cite{yang2025cogvideoxtexttovideodiffusionmodels})
& 0.9766 & 0.6520 & 0.3904 & 0.2616 \\

\midrule
\textbf{Ours} (CogVideoX-5B)
& \textbf{0.9831} & 0.6718 & \textbf{0.2604} & \textbf{0.4114} \\
\bottomrule
\end{tabular}
}
\end{table*}

\subsection{Localized Action Classification}
\label{sec:localized_action_retrieval}

A key property of localized motion representations is the ability to capture and semantically describe the motion of a subject of interest while remaining invariant to its appearance and unaffected by other entities in the scene. To evaluate this capability, we test whether motion embeddings extracted from a spatial query encode the underlying action dynamics independently of visual appearance and the actions performed by other actors.

\paragraph{Evaluation Setup}
We consider a \emph{zero-shot localized action classification} task: given a video and a spatial query specifying an actor, the goal is to classify the action performed by that actor using only the extracted motion embeddings. We construct this benchmark using the A2D dataset~\cite{xu2015can}, which consists of videos depicting multiple actors, each annotated with segmentation masks and an associated action label. For each actor--action instance, we extract a motion embedding conditioned on the corresponding segmentation mask and classify the action using a $k$-nearest neighbor (kNN) classifier. To ensure the evaluation emphasizes motion rather than appearance, instances of the same actor are excluded from the retrieval pool. Correct classification therefore requires identifying the underlying motion pattern rather than relying on actor identity or visual cues. We compare against image-, video-, and motion-based methods and report top-1 Accuracy and weighted F1 score to account for class imbalance. To adapt the other methods to this localized setting, we either crop the input video (${\textsc{rgb}}$), or the resulting spatial latent representations (${\textsc{latent}}$). If applicable, we report results for both strategies. We additionally construct and evaluate two constrained versions of our method, one where we apply the aforementioned cropping mechanism (\textbf{Ours}$_{\textsc{rgb}}$), and one where we do not apply localized querying at all (\textbf{Ours}$_{\textsc{global}}$). Results are summarized in \cref{tab:a2d_combined_results}.

\paragraph{Context is Important for Localized Motion Understanding} Image-based representations such as DINOv2~\cite{oquab_dinov2_2023} perform poorly on this task, highlighting the limitations of appearance features for localized action recognition. Video representations such as V-JEPA2~\cite{assran2025v} improve performance but remain constrained by their lack of explicit motion modeling. Motion-focused approaches such as DisMo~\cite{ressler-antal2025dismo} achieve stronger results, confirming the importance of motion representations for this task. 
Nonetheless, our method outperforms these baselines by a significant margin. While other methods lack access to the surrounding context due to the cropping mechanism, our model preserves the full scene context when conditioning the representation on the queried actor, allowing it to leverage contextual cues that help disambiguate motion without compromising spatial locality. We ablate the importance of context by providing our model with cropped input videos as well, effectively reducing it to the same context-reduced setting as the other methods (see \textbf{Ours}$_{\textsc{rgb}}$ in \cref{tab:a2d_combined_results}). Our method suffers from a substantial drop in performance, approaching the performance of other motion encoders in this setting, which suggests that contextual information plays a vital role for classifying localized actions.
However, providing global methods with scene context by processing uncropped videos is insufficient, as motion of other actors in the scene may compromise regional understanding. To highlight the influence of distractors, we reduce our method to this setting by prompting it with a global mask rather than actor-specific ones (see \textbf{Ours}$_{\textsc{global}}$ in \cref{tab:a2d_combined_results}). Once again, we notice a significant performance drop, indicating that either side of this trade-off is suboptimal. When not subjecting ourselves to either of these constraints, our approach performs significantly better, indicating that it does not sacrifice locality for context, but instead combines the strengths of both to enable improved action recognition.

\begin{table}[t]
\centering

\caption{\textbf{Localized Motion Classification.}
\textbf{Left:} Comparison of image, video, and motion encoders for localized action classification. Our method outperforms global approaches that obtain localization by cropping the video input (${\textsc{rgb}}$) or masking features in latent space (${\textsc{global}}$). Ablations show that removing context through RGB crops (\textbf{Ours}$_{\textsc{rgb}}$) or removing localization by querying the entire frame (\textbf{Ours}$_{\textsc{global}}$) both degrade performance. The best results are achieved when motion is encoded locally while preserving scene context. \textbf{Right:} Example of a positive and negative classification outcome with regard to a reference subject from the A2D dataset.}
\label{tab:a2d_combined_results}

\begin{tabular}{@{}cc@{}}
    {\adjustbox{valign=t, width=.50\linewidth}{\tablescalebox{
    \begin{tabular}{lcc}
    \toprule
     & \textbf{Accuracy}$\uparrow$ & \textbf{F1}$\uparrow$ \\
    \midrule
    
    \textit{\textbf{Image encoders}} \\
    
    DINOv2$_{\textsc{rgb}}$ \cite{oquab_dinov2_2023}      & 7.92 & 12.91 \\
    DINOv2$_{\textsc{latent}}$   & 1.90 & 3.43 \\
    
    \midrule
    
    \textit{\textbf{Video encoders}} \\
    
    V-JEPA2$_{\textsc{rgb}}$ \cite{assran2025v}     & 14.04 & 21.67 \\
    V-JEPA2$_{\textsc{latent}}$   & 12.04 & 18.65 \\
    
    \midrule
    
    \textit{\textbf{Global motion encoders}} \\
    
    DINOv2$_{\textsc{rgb}}$ + SemanticMoments \cite{huberman2026semanticmoments} & 11.93 & 18.28 \\
    DINOv2$_{\textsc{latent}}$ + SemanticMoments & 2.85 & 4.75 \\
    V-JEPA2$_{\textsc{rgb}}$ + SemanticMoments & 14.15 & 21.48 \\
    V-JEPA2$_{\textsc{latent}}$ + SemanticMoments & 12.99 & 18.76 \\
    DisMo$_{\textsc{rgb}}$ \cite{ressler-antal2025dismo}        & 27.24 & 38.96 \\
    \textbf{Ours}$_{\textsc{rgb}}$    & 27.98 & 39.21 \\
    \textbf{Ours}$_{\textsc{global}}$    & 21.75 & 31.83 \\
    
    \midrule
    
    \textbf{Ours}                 & \textbf{33.90} & \textbf{46.43} \\
    
    \bottomrule
    \end{tabular}
    }}} &
    {\adjustbox{valign=t}{\includegraphics[width=.24\linewidth]{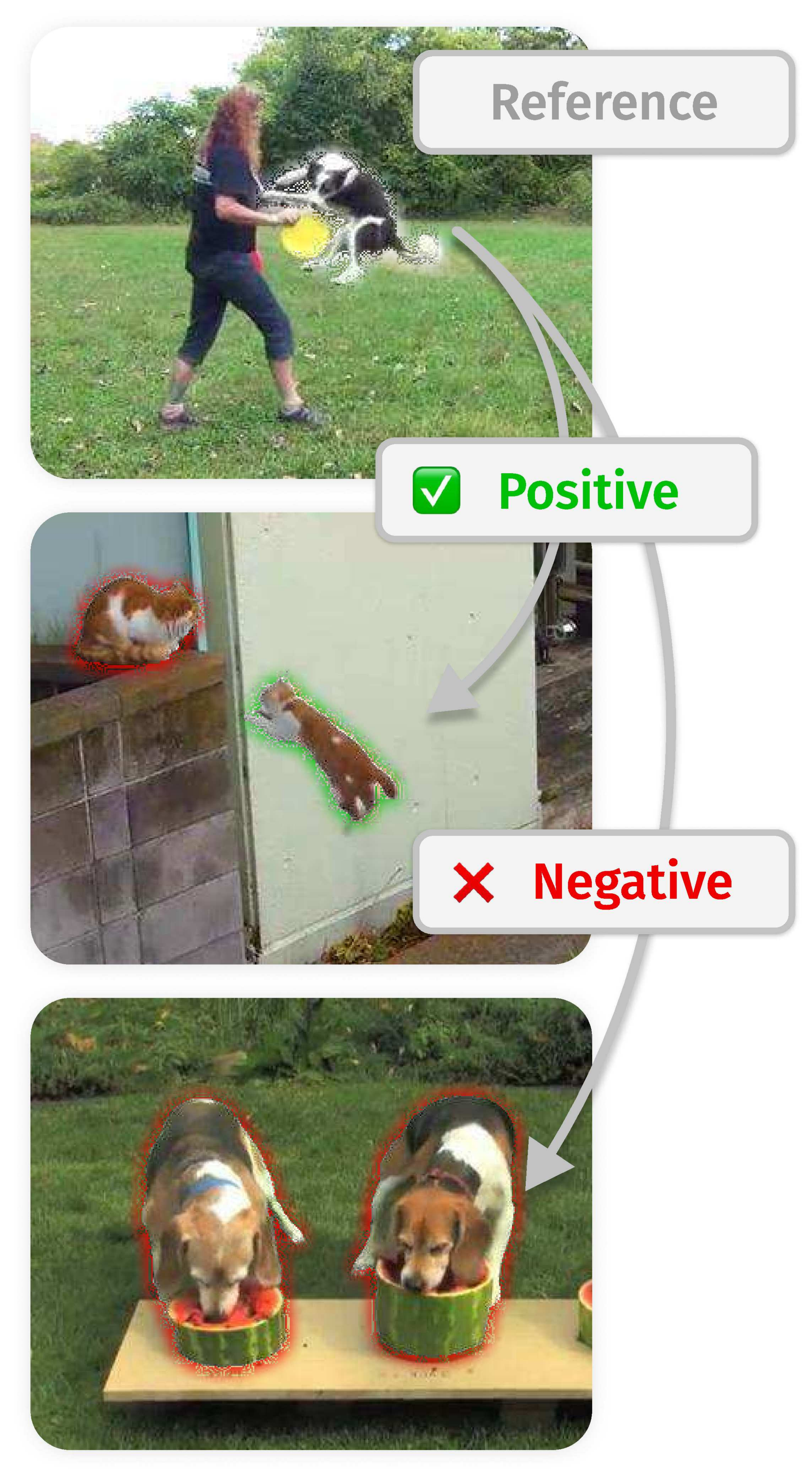}}}{\hskip -1em}
\end{tabular}
\end{table}

\subsection{From Local to Global: Generic Motion Representations}
\label{sec:global_experiments}

While our framework is designed to learn spatially localized motion representations, an important question is whether such representations remain useful beyond localized queries. Increasing spatial specificity could push representations toward low-level motion signals that lose higher-level semantic properties or overly emphasize local cues. In the following, we show that neither occurs: our learned embeddings generalize naturally to standard global motion understanding tasks while retaining the semantic properties characteristic of global motion representations.

\paragraph{Localized Motion Pre-training Generalizes to Global Understanding}
We evaluate our model in a \emph{global zero-shot action classification} setting on a diverse set of motion-centric video benchmarks. Specifically, we evaluate on ARID~\citep{xu2021arid}, which measures action recognition robustness under low-light conditions; IARD~\citep{DVN/DMT0PG_2019}, which evaluates identity-invariant motion understanding by minimizing reliance on appearance cues; Jester~\citep{materzynska2019jester}, a large-scale hand-gesture dataset that requires sensitivity to fine-grained local motions; Something-Something V2~\citep{goyal2017something}, which focuses on subtle human-object interactions and temporal reasoning; and Diving48~\citep{li2018resound}, a challenging fine-grained action recognition benchmark designed to reduce scene and object bias, requiring recognition based primarily on body dynamics and motion patterns. Although our framework is designed to learn \emph{localized} motion representations, these tasks require predicting a \emph{single label for the entire scene}. To adapt our method without architectural changes, we simply provide a mask covering the full start frame when computing the content query. This effectively reduces the localized motion encoder to a \emph{global motion encoder}, while leaving the remainder of the pipeline unchanged. The resulting performance is reported in \cref{tab:zs_action}. Our method achieves strong performance across all benchmarks and obtains the best results on several datasets. This demonstrates that localized motion pre-training does not limit the model to localized inference. Instead, the learned representation naturally extends to the global setting and remains highly effective for standard action recognition tasks.
Importantly, our training paradigm therefore yields a \emph{strictly more general motion encoder}. When no spatial conditioning is provided, the model behaves as a global motion encoder and achieves competitive or state-of-the-art performance. At the same time, it retains the ability to extract \emph{region-specific motion representations} when a query mask is given. In contrast to prior approaches that are inherently restricted to global motion encoding, our model seamlessly supports both regimes within the same representation.

\begin{table}[t]
\centering

\begin{minipage}[t]{0.63\linewidth}
\centering
\caption{\textbf{Zero-shot action classification in a global setting.}
We report top-1 accuracy on several motion-centric video benchmarks using a \textit{k}NN classifier with \textit{k=20}. Although our model is trained to produce localized motion representations, providing a full-frame query reduces it to a global motion encoder, yielding strong performance across benchmarks.}
\label{tab:zs_action}
\tablescalebox{
\begin{tabular}{lccccc}
\toprule
 & \textbf{SSv2}$\uparrow$ & \textbf{Jester}$\uparrow$ & \textbf{Diving48}$\uparrow$ & \textbf{ARID}$\uparrow$ & \textbf{IARD}$\uparrow$ \\
\midrule

DINOv2 \cite{oquab_dinov2_2023}     & 10.3 & 23.2 & 9.39 & 14.5 & 76.1 \\

\midrule

VideoMAE \cite{tong2022videomae} & 7.1 & 20.1 & - & 17.3 & 73.4 \\

V-JEPA2 \cite{assran2025v}          & 22.2 & 40.8 & 11.1 & 28.0 & 87.6 \\

\midrule

SemanticMoments$_{\text{DINO}}$ \cite{huberman2026semanticmoments}
                 & 11.4 & 37.3 & 9.85 & 22.0 & 62.7 \\
SemanticMoments$_{\text{V-JEPA2}}$ \cite{huberman2026semanticmoments}
                 & \textbf{31.5} & 52.2 & 12.7 & 38.3 & \underline{92.5} \\

DisMo \cite{ressler-antal2025dismo} & 24.6 & \underline{69.8} & \textbf{20.9} & \underline{55.3} & 92.0 \\

\midrule

\textbf{Ours} & \underline{26.5} & \textbf{72.8} & \underline{17.1} & \textbf{55.4} & \textbf{93.0} \\

\bottomrule
\end{tabular}
}
\end{minipage}
\hfill
\begin{minipage}[t]{0.325\linewidth}
\centering
\caption{\textbf{Motion retrieval under appearance transformations.} 
We report top-1 retrieval accuracy on the SemanticMoments synthetic split averaged over all augmentations.}
\label{tab:semantic_motion_results}
\tablescalebox{
\begin{tabular}{lc}
\toprule
 & \textbf{Accuracy}$\uparrow$ \\
\midrule

DINOv2 \cite{oquab_dinov2_2023}          & 44.4 \\

\midrule

VideoMAE \cite{tong2022videomae}         & 79.2 \\
V-JEPA2 \cite{assran2025v}         & 74.4 \\

\midrule

SemanticMoments$_{\text{DINO}}$  \cite{huberman2026semanticmoments}      & 86.4 \\
SemanticMoments$_{\text{V-JEPA2}}$ \cite{huberman2026semanticmoments}     & 84.4 \\

DisMo \cite{ressler-antal2025dismo}            & \textbf{90.0} \\

\midrule

\textbf{Ours}    & \underline{89.6} \\

\bottomrule
\end{tabular}
}
\end{minipage}

\end{table}

\paragraph{Spatial Localization Preserves Appearance Invariance} A key challenge for localized motion representations is that increasing spatial specificity can push representations toward low-level signals like optical flow, which capture motion locally but lose semantic invariances. Conversely, global video representations exhibit strong appearance invariance but cannot isolate individual entities' motion. Our goal is to combine both advantages: spatially localized representations that preserve the semantic invariances of global ones. Specifically, we target invariance to appearance-driven transformations, namely viewpoint changes, dynamic attribute modifications, entity replacement across semantic classes, and stylistic rendering changes, since these are critical for transferability to novel structures.\\

\begin{wrapfigure}{r}{0.3\linewidth}
    \vspace{-2em}
    \centering
    \includegraphics[width=0.9\linewidth]{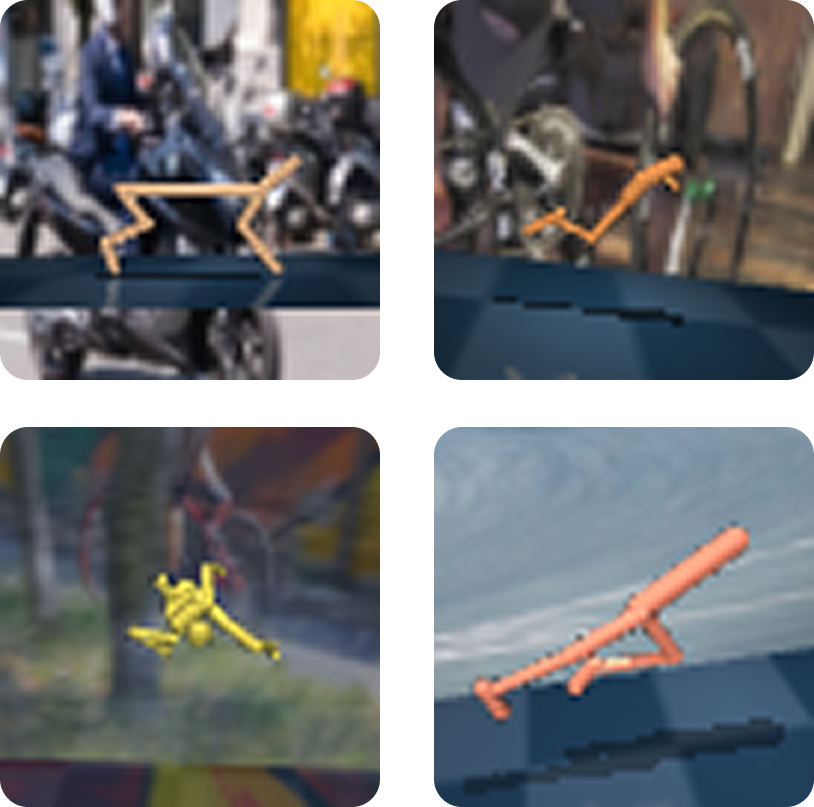}
    \caption{\textbf{DCS examples.} The dataset introduces visual distractors through randomized agent colors, video backgrounds, and camera motion.}
    \label{fig:dcs_examples}
    \vspace{-3em}
\end{wrapfigure}
We evaluate on the synthetic split of the \textit{SemanticMoments} benchmark\cite{huberman2026semanticmoments}, which measures motion retrieval under controlled appearance transformations (added objects, appearance changes, object replacement, viewpoint variation, style changes) using a top-1 retrieval protocol. We compare against DINOv2\cite{oquab_dinov2_2023}, V-JEPA2\cite{assran2025v}, and DisMo\cite{ressler-antal2025dismo}, applying the benchmark's temporal aggregation to DINOv2 and V-JEPA2. Results  are shown in \cref{tab:semantic_motion_results}. These show our method outperforms all baselines except DisMo, achieving near-parity while substantially surpassing appearance-centric representations. This demonstrates that spatial localization does not push representations toward low-level motion cues; instead, it preserves semantically meaningful motion dynamics while enabling localized reasoning.

\label{sec:dcs_probe}
\paragraph{Action Probing under Distractors} Following MaskLAM~\cite{fechner2026segment}, we test action alignment under visual distractors using the public LAOM DCS dataset~\cite{Stone2021TheDC,TUNYASUVUNAKOOL2020100022} (for examples see \cref{fig:dcs_examples}). We freeze each representation and train only a linear action probe to predict the actions. Our localized LAM obtains the lowest mean NMSE, improving over uncropped DisMo by 32.3\%, while direct RGB masking also substantially improves DisMo but remains weaker than our representation. These results show that spatial localization improves the robustness and action alignment of motion representations under visual distractors.

\begin{table}[h]
\centering
\caption{\textbf{Frozen action probing under visual distractors.}
We report normalized MSE (NMSE; lower is better) on the public LAOM DCS dataset using a linear action probe on frozen representations. Localizing the motion representation substantially improves over global DisMo and also outperforms RGB cropping, indicating that representation-level localization better preserves action-relevant information under distractors. \textcolor{gray}{MaskLAM} results are taken from~\cite{fechner2026segment} and use a larger training dataset.}

\label{tab:dcs_probe}
\tablescalebox{
\begin{tabular}{lccccc}
\toprule
 & \multicolumn{5}{c}{\textbf{NMSE} $\downarrow$} \\
\cmidrule(lr){2-6}
 & \textbf{Cheetah Run} & \textbf{Hopper Hop} & \textbf{Humanoid Walk} & \textbf{Walker Run} & \textbf{Mean} \\
\midrule
\textcolor{gray}{MaskLAM~\cite{fechner2026segment}} & \textcolor{gray}{0.1148} & \textcolor{gray}{0.2601} & \textcolor{gray}{0.3852} & \textcolor{gray}{0.1704} & \textcolor{gray}{0.2326} \\
\midrule
DisMo~\cite{ressler-antal2025dismo} & 0.6095 & 0.7250 & 0.6870 & 0.5446 & 0.6415 \\
DisMo (RGB cropped)~\cite{ressler-antal2025dismo} & 0.4201 & 0.5406 & 0.5225 & 0.5138 & 0.4993 \\
\textbf{Ours} & \textbf{0.4161} & \textbf{0.4744} & \textbf{0.4605} & \textbf{0.3845} & \textbf{0.4339} \\
\bottomrule
\vspace{-3em}
\end{tabular}
}
\end{table}

%% file: sec/5_conclusion.tex
\section{Conclusion}
\label{sec:conclusion}
In this work, we introduce a promptable and localized motion representation that encodes region-specific dynamics while preserving full-scene context. We showed that local motion cannot be reliably inferred from spatial crops, which discard context, nor from post-hoc feature masking, which does not condition the encoding process on the queried region. By conditioning motion encoding directly on spatial masks while processing the full video, our approach produces temporally consistent and spatially precise motion embeddings. Across motion transfer and localized classification tasks, we demonstrated improved controllability, structured scene composition, and stronger motion understanding compared to global video representation localized via cropping or feature masking. We achieve these results, while still demonstrating strong performance on global video understanding tasks, such as classification and retrieval. Our results highlight the importance of encoding local motion while preserving full-scene context for structured and compositional video modeling.

\section*{Acknowledgment}
This project has been supported by the Horizon Europe project ELLIOT (GA No.\ 101214398), the project ``GeniusRobot'' (01IS24083) funded by the Federal Ministry of Research, Technology and Space (BMFTR), the BMWE ZIM-project (No.\ KK5785001LO4) ``conIDitional LoRA'', the German Federal Ministry for Economic Affairs and Energy within the project ``NXT GEN AI METHODS - Generative Methoden für Perzeption, Prädiktion und Planung'', the bidt project KLIMA-MEMES, and the German Academic Exchange Service (DAAD) under the Kondrad Zuse School of Excellence for Reliable AI (RelAI). The authors gratefully acknowledge the Gauss Center for Supercomputing for providing compute through the NIC on JUWELS/JUPITER at JSC and the HPC resources supplied by the NHR@FAU Erlangen. We thank Timy Phan and for their helpful feedback and support, and Owen Vincent for continuous technical support.

%% file: sec/X_supp.tex
\clearpage
\hypersetup{pageanchor=false}
\setcounter{page}{1}

\setcounter{figure}{0}
\setcounter{table}{0}
\setcounter{equation}{0}
\setcounter{section}{0}
\renewcommand\thesection{\Alph{section}}
\renewcommand\thefigure{\Alph{section}.\arabic{figure}}
\renewcommand\thetable{\Alph{section}.\arabic{table}}
\renewcommand\theequation{\Alph{section}.\arabic{equation}}

\renewcommand\theHsection{appsec.\Alph{section}}
\renewcommand\theHsubsection{appsec.\Alph{section}.\arabic{subsection}}
\renewcommand\theHfigure{appfig.\Alph{section}.\arabic{figure}}
\renewcommand\theHtable{apptab.\Alph{section}.\arabic{table}}
\renewcommand\theHequation{appeq.\Alph{section}.\arabic{equation}}

\crefalias{section}{appsec}
\crefalias{figure}{appfig}
\crefalias{table}{apptab}
\crefalias{equation}{appeq}

\section*{Supplementary Material}

\section{Implementation Details}\label{sec:app_implementation_details}

\subsection{Pretraining details}

\begin{table}[t]
\centering
\caption{\textbf{Pretraining hyperparameters for our model.} Summary of training data, preprocessing, optimization settings, and architecture configuration.}
\label{tab:implementation_details}
\tablescalebox{
\begin{tabular}{lll}
\toprule
\textbf{Category} & \textbf{Component} & \textbf{Setting} \\
\midrule

\multirow{4}{*}{Data}
& Frames per video & 8 \\
& Frame resolution & $256 \times 256$ \\
& Tracks per decoding step & $256$ \\
& Masks per video & $4$ \\

\midrule

\multirow{4}{*}{Transformations}
& Random resized crop & ($0.04$, $1.0$) \\
& Rotation & ($-30$, $30$) \\
& Aspect ratio & ($3/4$, $4/3$) \\
& Brightness & ($0.5$, $1.5$) \\
& Contrast & ($0.5$, $1.5$) \\
& Saturation & ($0.5$, $1.5$) \\

\midrule

\multirow{5}{*}{Optimization}
& Batch size & $64$ \\
& Training schedule & 600k iterations \\
& Optimizer & AdamW \\
& Warmup iterations & 5000 \\
& lr scheduler & constant w/ warmup \\
& lr & $5 \times 5^{-5}$ \\
& Weight decay & $0.01$ \\

\midrule

\multirow{10}{*}{Architecture}
& Frame embedder & ViT-Base (12 layers, dim 768) \\
& Frame embedder init & DINOv2 pretrained weights \\
& Content encoder & 4-layer cross-attention aggregator \\
& Content query tokens & 1 per entity \\
& Content embedding dim & 768 \\
& Motion encoder & 12-layer Vision Transformer \\
& Motion queries & Content embeddings repeated across frames \\
& Tracks decoder & 12-layer Vision Transformer \\
& Tracks decoder conditioning & Adaptive RMSNorm (motion + content) \\

\bottomrule
\end{tabular}
}
\end{table}

Our architecture consists of four main components: a \emph{frame embedder}, a \emph{content encoder}, a \emph{motion encoder}, and a \emph{decoder}. All transformer-based modules operate with a hidden dimension of $768$. Details about training data, optimization, and method architecture can be found in ~\cref{tab:implementation_details}.

\paragraph{Frame Embedder}
Input video frames are first processed by a shared frame embedding network that produces per-frame token embeddings. The frame embedder is implemented as a $12$-layer Vision Transformer with hidden dimension $768$. To accelerate convergence and leverage strong visual representations, the model is initialized with weights from \cite{oquab_dinov2_2023}. The resulting frame tokens serve as input to all subsequent modules.

\paragraph{Content Encoder}
The content encoder provides an entity-specific appearance representation that identifies the region whose motion should be modeled. Given a video frame and a corresponding entity mask, it acts as a mask-conditioned region encoder that maps the visual content of the selected entity to a single content embedding. This formulation is independent of the particular aggregation architecture: the region encoder can, for example, be implemented through masked pooling, mask-conditioned attention over image tokens, or an image encoder that receives the mask as an additional conditioning signal. For the experiments in this work, we instantiate the content encoder as a masked attentive pooling network over spatial frame embeddings. A learnable query token attends to the frame embeddings through four cross-attention layers, where the entity mask is used to bias the aggregation toward the target region. Each cross-attention layer is followed by an MLP block for further feature aggregation. The resulting query token forms the \emph{content embedding} of the entity.

\paragraph{Motion Encoder}
The motion encoder produces motion embeddings for each entity across the temporal dimension of the video. It takes as input the frame embeddings of all frames together with the entity-specific content embeddings.
For each entity, the corresponding content embedding is repeated across the temporal dimension and used as a query token. These queries interact with the frame embeddings through a $12$-layer Vision Transformer with hidden dimension $768$, resulting in \emph{motion embeddings} that describe the entity-specific motion at each frame.

\paragraph{Decoder}
The decoder predicts motion in the form of point trajectories relative to a source frame. For each entity, we sample a set of starting points within the corresponding entity region of the source frame and obtain their ground-truth trajectories across the video using TAPNext~\citep{tapnext}. The decoder is then trained to predict these temporal trajectories for a fixed number of prediction steps.
The decoder receives the frame embeddings of the source frame together with the corresponding motion and content embeddings. It is implemented as a $12$-layer Vision Transformer with hidden dimension $768$. Conditioning on motion and content embeddings is implemented using \emph{adaptive RMSNorm}, allowing the decoder activations to be modulated by the motion and content signals. The decoder outputs the predicted trajectories for the sampled points, expressed as displacements relative to their initial positions in the source frame.

\paragraph{Training Data}
We train on a mixture of two datasets covering approximately 17M video clips. The first one is OpenVid-1M~\citep{nan2025openvidm}, a high-quality open-domain dataset well suited for generative tasks. The second one is a self-collected dataset of approximately 16M clips sourced from YouTube, covering a broad range of real-world scenes and motions. We provide an overview of the dataset composition in \cref{tab:training_dataset_composition}.

\begin{table}[htb]
    \centering
    \caption{Training data composition.}
    \adjustbox{max width=\linewidth}{
        \begin{tabular}{lcc}
            \toprule
            Dataset & Domain & \# Clips \\
            \midrule
            OpenVid-1M~\citep{nan2025openvidm} & open & 1M \\
            Internal & open & 16M \\
            \midrule
            \textit{Combined} & open & 17M \\
            \bottomrule
        \end{tabular}
    }
    \label{tab:training_dataset_composition}
\end{table}

\subsection{Video Model Finetuning Details}

\paragraph{Training Setup}
We adapt the attention and feedforward layers of CogVideoX-5B~\cite{yang2025cogvideoxtexttovideodiffusionmodels} using LoRA~\citep{hu2021lora} modules with rank $128$. We fine-tune the model using a batch size of $16$ and the AdamW optimizer with $\beta_1 = 0.9$, $\beta_2 = 0.95$ and learning rate $5 \times 10^{-5}$ with a constant schedule and $250$ linear warmup steps. Training is performed in \texttt{bfloat16} mixed precision. Videos are processed at a resolution of $25 \times 480 \times 720$ (frames $\times$ height $\times$ width) and generated at $8$~FPS.

\paragraph{Temporal Alignment of Motion Conditioning}
The CogVideoX VAE compresses $T{=}25$ input frames into $N_\ell{=}7$ latent frames using a temporal compression ratio of $c_t{=}4$. Our motion representation model extracts embeddings via a sliding window of size $p{+}1$ with prediction horizon $p{=}3$, producing $T - (p{+}1) = 21$ motion embeddings per region. To align these with the VAE latent frames, we prepend $p = 3$ learned padding tokens, yielding $T - 1 = 24$ temporally contiguous embeddings. These are grouped into non-overlapping chunks of $c_t{=}4$, producing $(T{-}1)/c_t = 6$ temporally aligned representations. Each chunk is concatenated along the feature dimension and projected through a learned mapping network to produce a single conditioning vector per latent frame. A final learned padding vector is appended for the last latent frame, yielding $N_\ell{=}7$ conditioning vectors in one-to-one correspondence with the VAE latent frames. Motion and content tokens are concatenated along the channel dimension before this projection, allowing the model to condition on joint motion--content representations while keeping the base weights frozen.

\paragraph{Multi-region Conditioning}
For multi-region scenarios, the above alignment procedure is applied independently per region mask, producing $K \times N_\ell$ conditioning tokens for $K$ regions. All per-region tokens are concatenated along the sequence dimension and appended to the self-attention layers of the transformer. To preserve spatial and temporal grounding, we apply 3D rotary positional embeddings (RoPE)~\citep{su2023roformerenhancedtransformerrotary}  to all conditioning tokens. The temporal position of each token is determined by its corresponding latent frame index, while the spatial position is set to the centroid of its associated region mask. This ensures that each motion token attends with positional bias consistent with the spatiotemporal location of its corresponding region in the generated video.

\subsection{Evaluation Details}
\paragraph{Datasets} We evaluate on a diverse set of benchmarks spanning action classification and retrieval, each emphasizing different aspects of temporal and spatial reasoning. Something-Something v2 (SSv2)~\citep{goyal2017something} is a large-scale dataset of human--object interactions designed to test fine-grained temporal reasoning while minimizing reliance on background context or object identity.
Jester~\citep{materzynska2019jester} contains over 148,000 short clips of hand gestures, serving as a benchmark for fine-grained, spatially localized motion sensitivity.
ARID~\citep{xu2021arid} (Action Recognition in the Dark) comprises 5,500 RGB-D videos captured under low-light conditions, evaluating robustness to appearance degradation.
IARD~\citep{DVN/DMT0PG_2019} (Invariant Action Recognition Dataset) consists of controlled recordings of five actors performing five actions from multiple viewpoints, enabling evaluation of identity- and viewpoint-invariant motion recognition.
Diving48~\citep{diving48} contains approximately 18,000 trimmed clips of competitive diving covering 48 standardized dive sequences, requiring modeling of long-term temporal dynamics across takeoff, flight, and entry stages.
A2D~\citep{xu2017actionunderstandingmultipleclasses} provides 3,782 videos with pixel-level actor and action annotations, enabling evaluation of spatially grounded action understanding across multiple actor--action combinations. The SemanticMoments synthetic benchmark~\citep{huberman2026semanticmoments} provides a controlled evaluation setting designed to isolate specific failure modes of video representations. By holding the underlying motion constant while systematically varying non-motion factors such as appearance, object identity, and viewpoint, it enables precise analysis of how these confounds influence representation similarity.

\paragraph{Motion Transfer using Track-conditioned Models}
To perform motion transfer using track-conditioned video generation baselines, we re-target point trajectories from a source video onto a target image before conditioning the generator. Given a source video with associated object masks, we first extract dense point tracks within the masked region using CoTracker3~\citep{cotracker3}. These tracks are then spatially transferred to align with the target objects by computing an affine mapping between the bounding boxes of the source and target masks: each track point is translated so that the source mask centroid coincides with the target mask centroid, and uniformly scaled so that the source bounding box fits within the target bounding box. The retargeted tracks, together with the target image, are then provided as conditioning inputs to the track-conditioned video generation model. 

\section{Ablations}
\label{sec:appendix_ablations}

In this section we present ablation studies that analyze several architectural and objective design choices of our model. These experiments aim to justify the configuration used in the final model and provide insight into how different components affect the learned motion representations.

Specifically, we investigate three aspects: 
(i) the prediction distance used during trajectory decoding, 
(ii) the dimensionality of the motion embedding bottleneck, and 
(iii) the contribution of the content encoder and its conditioning during trajectory decoding.

\subsection{Prediction Distance}
\label{sec:prediction_distance_ablation}

\begin{figure}[t]
    \centering
    \includegraphics[width=0.55\linewidth]{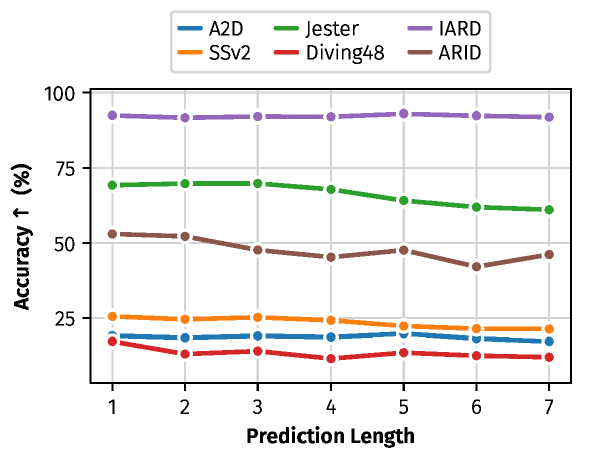}
    \caption{\textbf{Action classification accuracy across varying prediction lengths $p$.} Performance remains stable across $p \in \{1, \dots, 7\}$, with minor degradation beyond $p=3$, suggesting that our framework is robust to different choices of prediction length.}
    \label{fig:app_prediction_distance_ablation}
\end{figure}

We study the effect of the \emph{prediction distance} $p$ used during trajectory decoding, which determines how many steps into the future point trajectories are being predicted relative to their starting positions in the source frame. At one extreme, $p=1$ reduces to sparse optical flow estimation; at the other, $p=7$ requires predicting trajectories across the entire video (8 frames total). We train seven model variants with $p \in \{1,\dots,7\}$, each for 300k steps, and evaluate on six action classification benchmarks: A2D, Something-Something-V2 (SSv2), Jester, Diving48, IARD, and ARID. A2D is a localized multi-label benchmark whereas the remaining ones are global datasets with a single label per video clip. Results are summarized in \cref{fig:app_prediction_distance_ablation}.

Overall, the results indicate that our method is robust to the choice of prediction distance. While SSv2 and Jester exhibit a moderate performance drop beyond $p=3$, the differences remain small, suggesting that the prediction distance is not a critical design choice. We adopt $p=3$ as a practical default based on these results, though nearby values yield comparable performance.

\subsection{Motion Bottleneck Size}
\label{sec:motion_bottleneck_ablation}

\begin{figure}[t]
    \centering
    \includegraphics[width=0.55\linewidth]{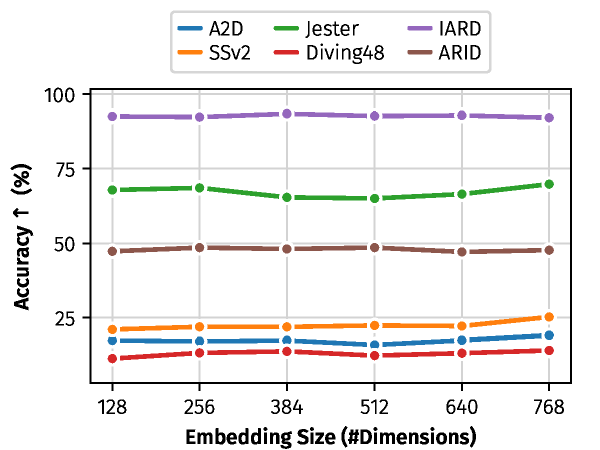}
    \caption{\textbf{Action classification accuracy across motion bottleneck sizes.} Performance remains stable from $d=768$ down to $d=128$, indicating that motion dynamics are effectively captured even with compact embeddings.}
    \label{fig:app_bottleneck_ablation}
\end{figure}

We study the effect of the motion embedding dimensionality on downstream performance. We train model variants with bottleneck sizes ranging from $d=128$ to $d=768$ and evaluate on the same action classification benchmarks used in \cref{sec:prediction_distance_ablation}. We adopt $d=768$ as the default in all main experiments. Results are summarized in \cref{fig:app_bottleneck_ablation}.\\

Across all benchmarks, performance remains remarkably stable even as the bottleneck size is reduced by a factor of six, from $d=768$ down to $d=128$. This indicates that the motion dynamics captured by our representation can be effectively compressed into compact embeddings without significant information loss. The robustness to bottleneck size suggests that even more aggressive compression may be feasible, making the approach well suited for downstream applications where compact representations are desirable.

\subsection{Appearance Entanglement Analysis}
\label{sec:app_attachment_entanglement}

\begin{figure}[t]
    \centering
    \includegraphics[width=0.55\linewidth]{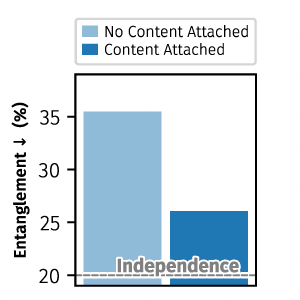}
    \caption{\textbf{Appearance entanglement analysis.}
    We quantify appearance leakage in motion embeddings on IARD by measuring actor identity classification using a $k$NN classifier. With five actors, perfect independence corresponds to $20\%$ accuracy. \textbf{Left:} Model without decoder conditioning on the content token. \textbf{Right:} Model with content-token conditioning. Conditioning markedly reduces identity predictability toward the independence line, highlighting the role of content attachment in disentangling motion from appearance.}
    \label{fig:attachment_ablation}
\end{figure}

We analyze the degree to which the learned motion embeddings remain entangled with appearance information. 
Ideally, motion representations should encode \emph{how} an entity moves while remaining independent of \emph{who} performs the motion.\\

To quantify appearance leakage, we evaluate motion embeddings on the IARD dataset by measuring actor identity classification with a $k$NN classifier. 
The dataset contains five actors; therefore, a representation that is fully independent of appearance should achieve approximately $20\%$ accuracy (chance level). 
Higher accuracy indicates that identity information remains encoded in the motion embedding. We compare two model variants that differ in how the decoder is conditioned during track reconstruction. 
Importantly, both variants already receive implicit appearance context: tracks are only queried at spatial locations that lie \emph{inside the masked regions}. 
Consequently, the decoder is naturally conditioned on the localized region of the entity, which already encourages a degree of appearance disentanglement. The difference between the two variants lies in whether the decoder additionally receives the corresponding \emph{content token}.  In the \emph{No Content Attachment} variant, reconstruction relies solely on the motion embedding and the spatial mask constraint. 
In the \emph{Content Attachment} variant, the decoder is additionally conditioned on the content token, providing a dedicated pathway for appearance information.\\

Figure~\ref{fig:attachment_ablation} shows the resulting actor identity classification accuracy of the motion embeddings. Even without explicit content attachment, identity predictability is already reduced due to the spatial mask conditioning. However, providing the content token during decoding further decreases identity classification accuracy, pushing performance closer to the independence line. 
This demonstrates that explicit content attachment strengthens the separation between motion and appearance, encouraging motion embeddings that are more appearance-invariant.

\begin{figure}[t]
\centering
\includegraphics[width=0.5\linewidth]{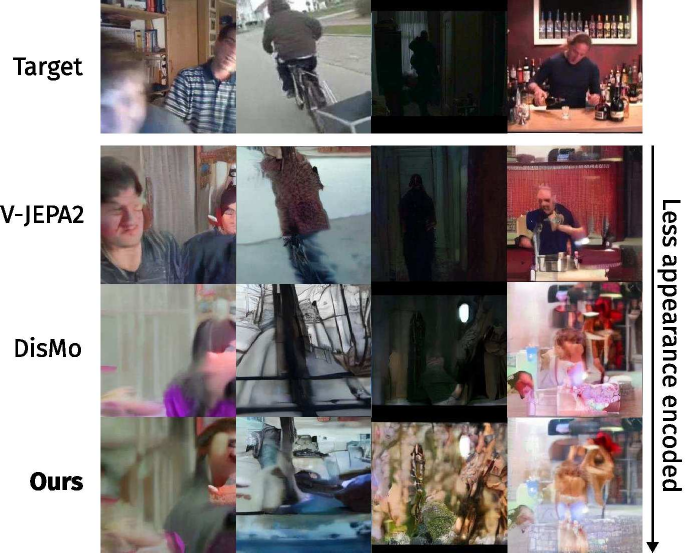}
\caption{\textbf{Appearance reconstruction probe examples.}
Reconstructions from our motion embeddings preserve less of the actor's appearance than those obtained from the baseline representations.}
\label{fig:decoder_probe}
\end{figure}

To further assess appearance leakage, we train a flow-matching decoder to reconstruct the input frame from frozen motion embeddings. Since the encoder cannot adapt during probe training, better reconstruction indicates that more appearance information is retained in its representation. As shown in \cref{tab:decoder_probe,fig:decoder_probe}, our embeddings yield the highest L1 and LPIPS errors and the lowest CLIP similarity, indicating that they contain less recoverable appearance information than the baseline representations.

\begin{figure}[t]
\centering
\captionof{table}{\textbf{Appearance reconstruction probe.} Lower CLIP similarity and higher L1 and LPIPS errors indicate less recoverable appearance information.}
\label{tab:decoder_probe}
\setlength{\tabcolsep}{4pt}
\resizebox{0.5\linewidth}{!}{%
\begin{tabular}{lccc}
\toprule
\textbf{Model} 
& \textbf{CLIP $\downarrow$} 
& \textbf{L1 $\uparrow$} 
& \textbf{LPIPS $\uparrow$} \\
\midrule
V-JEPA2 [28]
& 0.56 
& 0.19 
& 0.61 \\

DisMo [32]
& \textbf{0.53} 
& 0.26 
& 0.67 \\

Ours 
& \textbf{0.53}
& \textbf{0.27} 
& \textbf{0.69} \\
\bottomrule
\end{tabular}
}
\end{figure}

\section{Qualitative Examples}

We present additional qualitative results for motion transfer and compare them against the previously introduced baselines. The results cover both one-to-one transfer and compositional settings with multiple motion sources mapped to multiple target objects. All experiments are conducted using our adapted CogVideoX-5B model.
\subsection{Localized Motion Transfer}
\begin{figure*}[t]
    \centering
    \includegraphics[width=\linewidth]{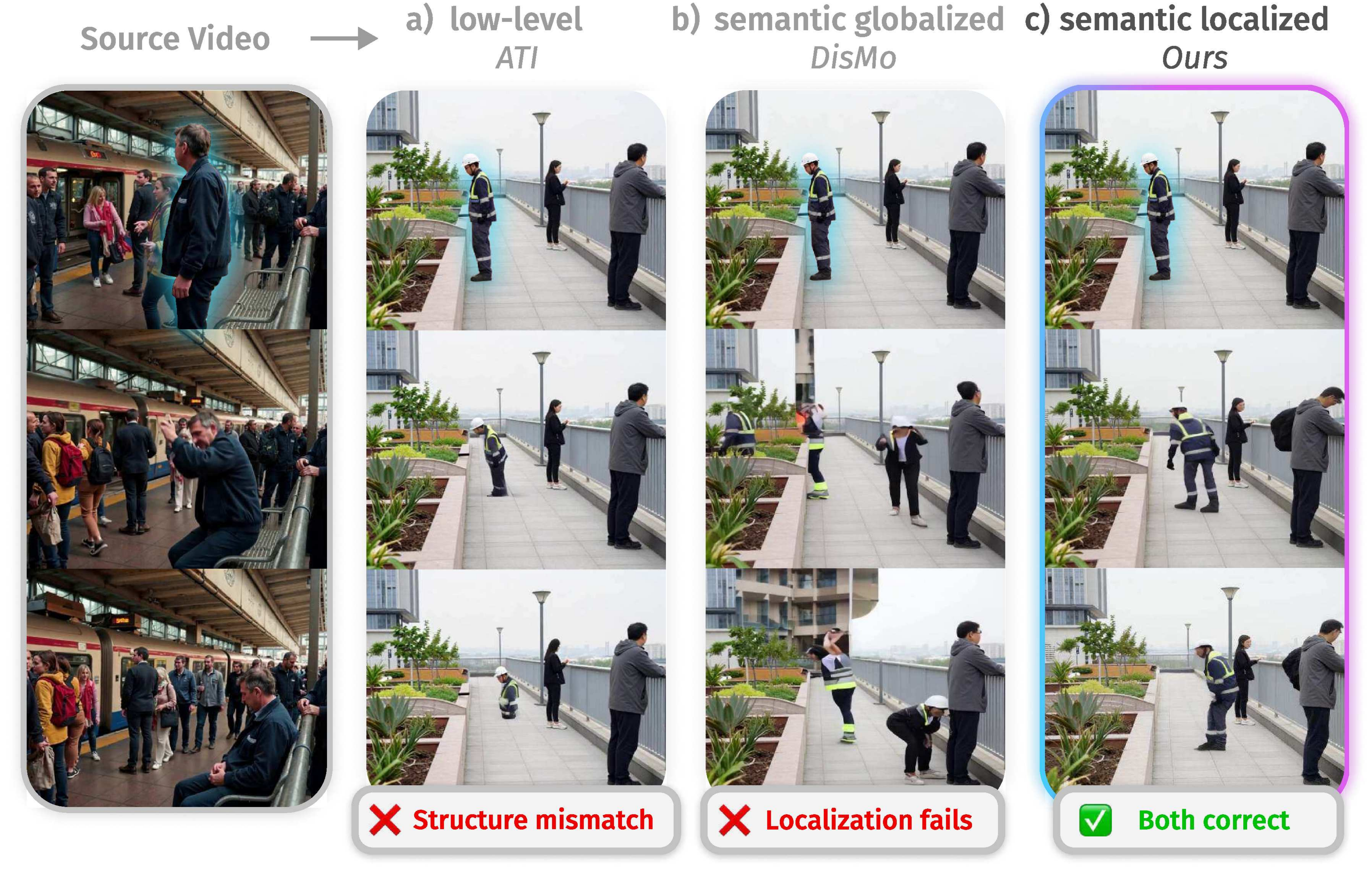}
    \caption{\textbf{Localized Motion Transfer.}
    Poses don't align, so instead of making the person sit, \textbf{(a)} makes the person clip through the ground.
    \textbf{(b)} moves the wrong person, \textbf{(c)} makes the correct person sit.
    }
    \label{fig:comp1}
\end{figure*}

\begin{figure*}[t]
    \centering
    \includegraphics[width=\linewidth]{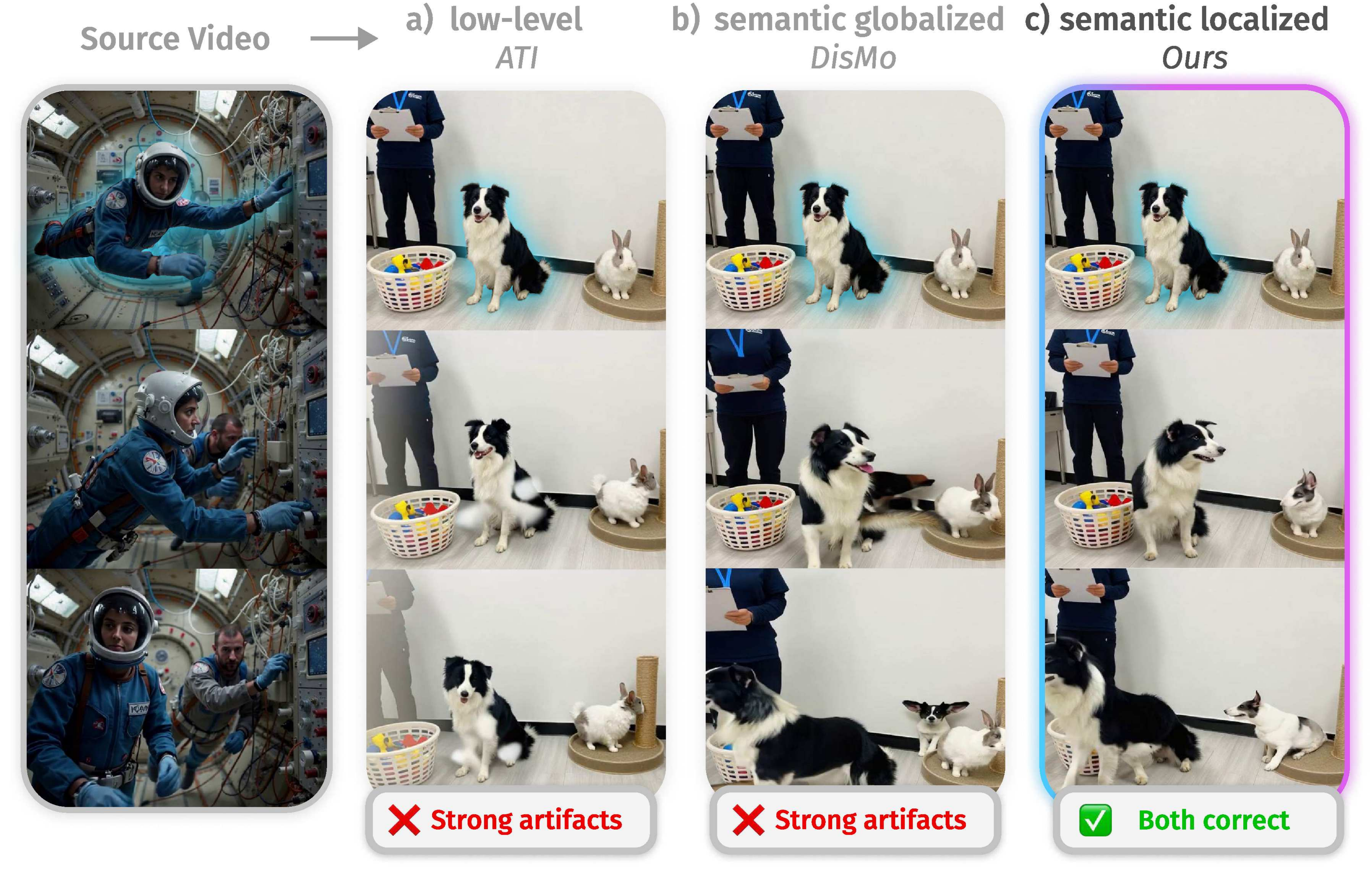}
    \caption{\textbf{Localized Motion Transfer.}
    Since the poses don't align, \textbf{(a)} generates strong artifacts, while \textbf{(b)} generates a second dog to accommodate for the two subjects in the source video, while
    \textbf{(c)} only moves the correct subject without strong artifacts.
    }
    \label{fig:comp3}
\end{figure*}

\begin{figure*}[t]
    \centering
    \includegraphics[width=\linewidth]{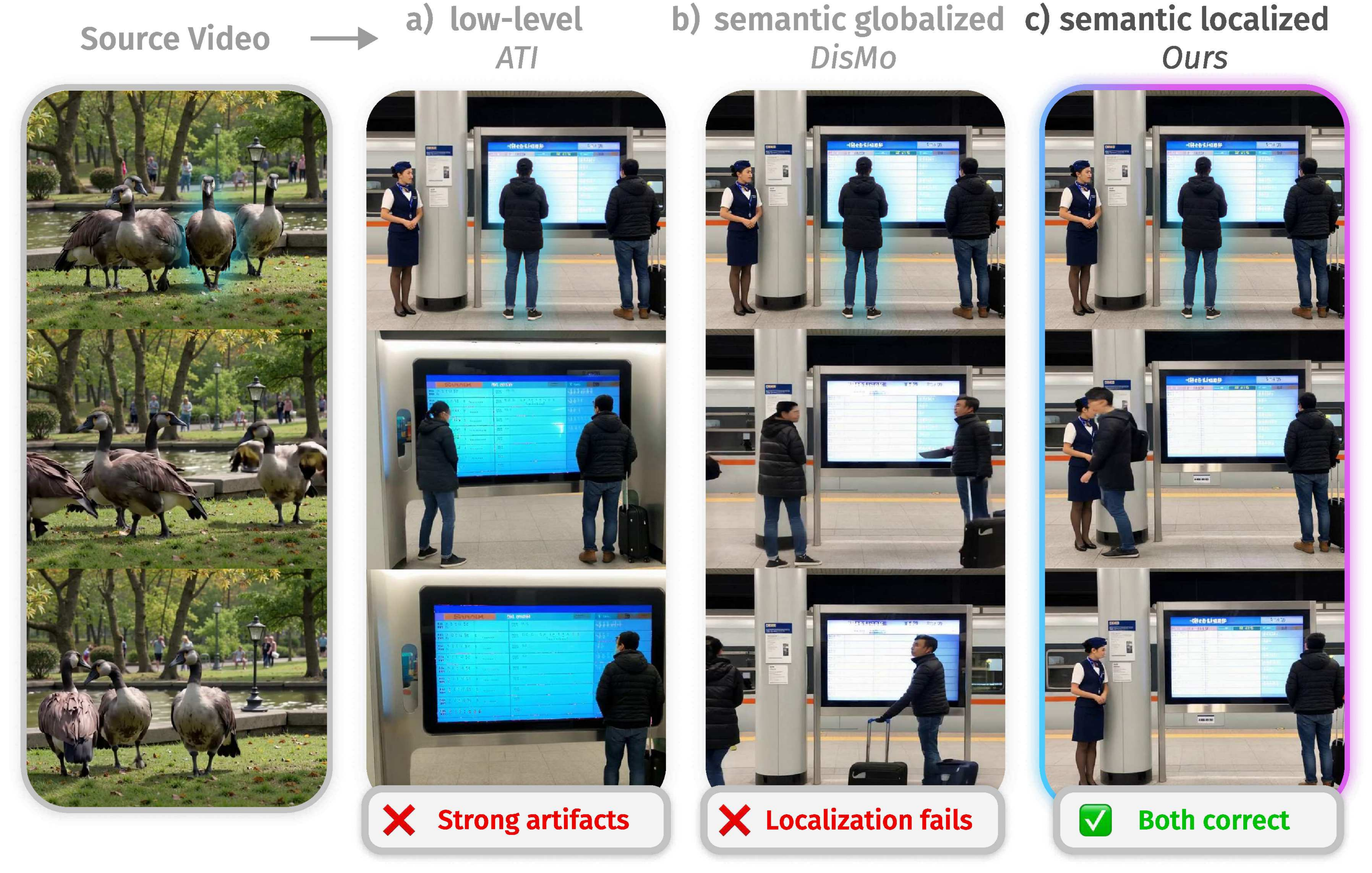}
    \caption{\textbf{Localized Motion Transfer.}
    Since the poses don't align, \textbf{(a)} generates strong artifacts, while \textbf{(b)} moves all subjects, however \textbf{(c)} moves only the correct person.
    }
    \label{fig:comp4}
\end{figure*}

\begin{figure*}[t]
    \centering
    \includegraphics[width=\linewidth]{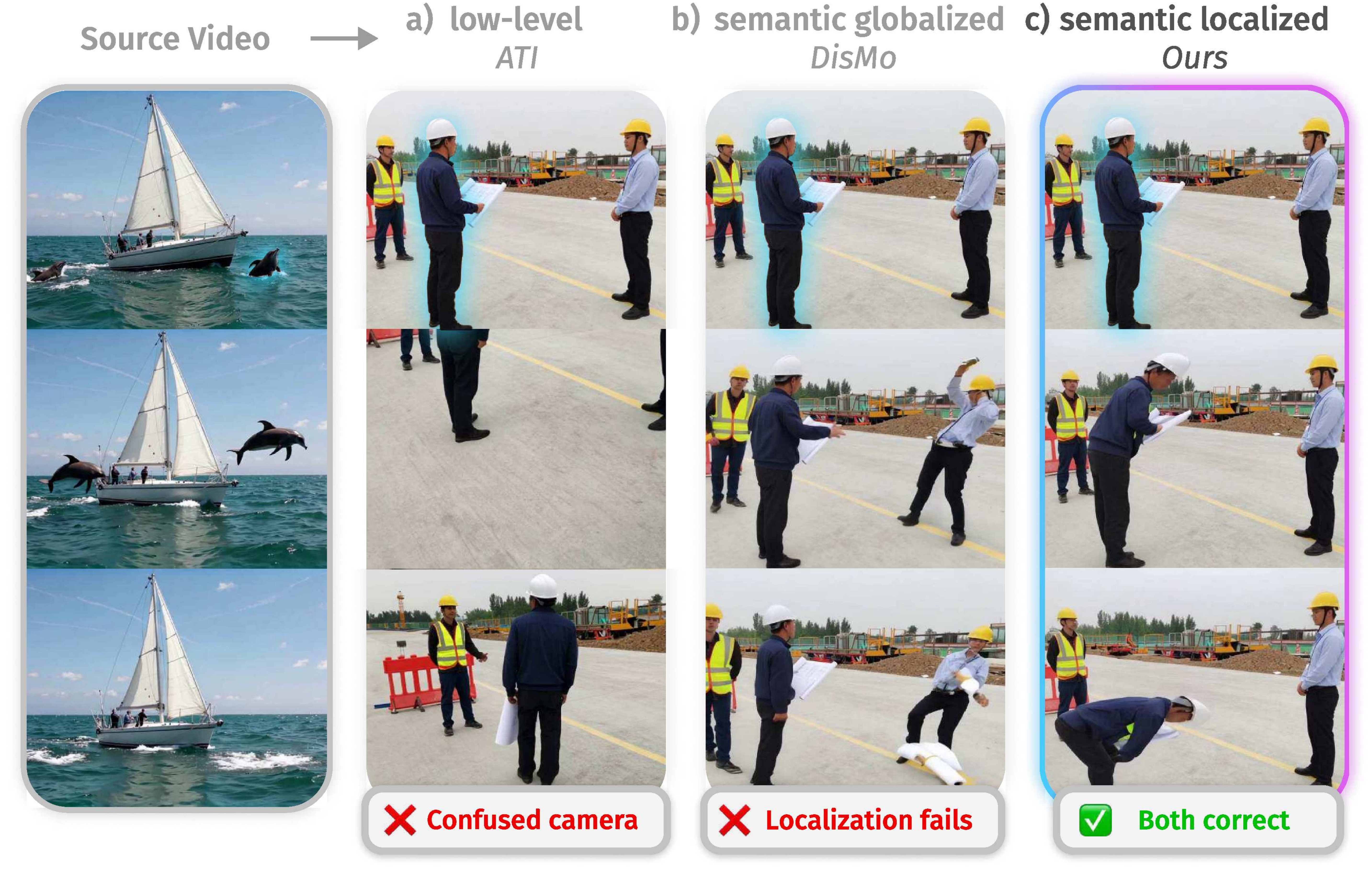}
    \caption{\textbf{Localized Motion Transfer.}
    \textbf{(a)} Confuses the object motion for camera motion and does not generate the correct motion since the poses don't align,
    \textbf{(b)} moves the wrong person with the wrong motion,
    \textbf{(c)} moves the correct person with the right motion.
    }
    \label{fig:comp5}
\end{figure*}

\begin{figure*}[t]
    \centering
    \includegraphics[width=\linewidth]{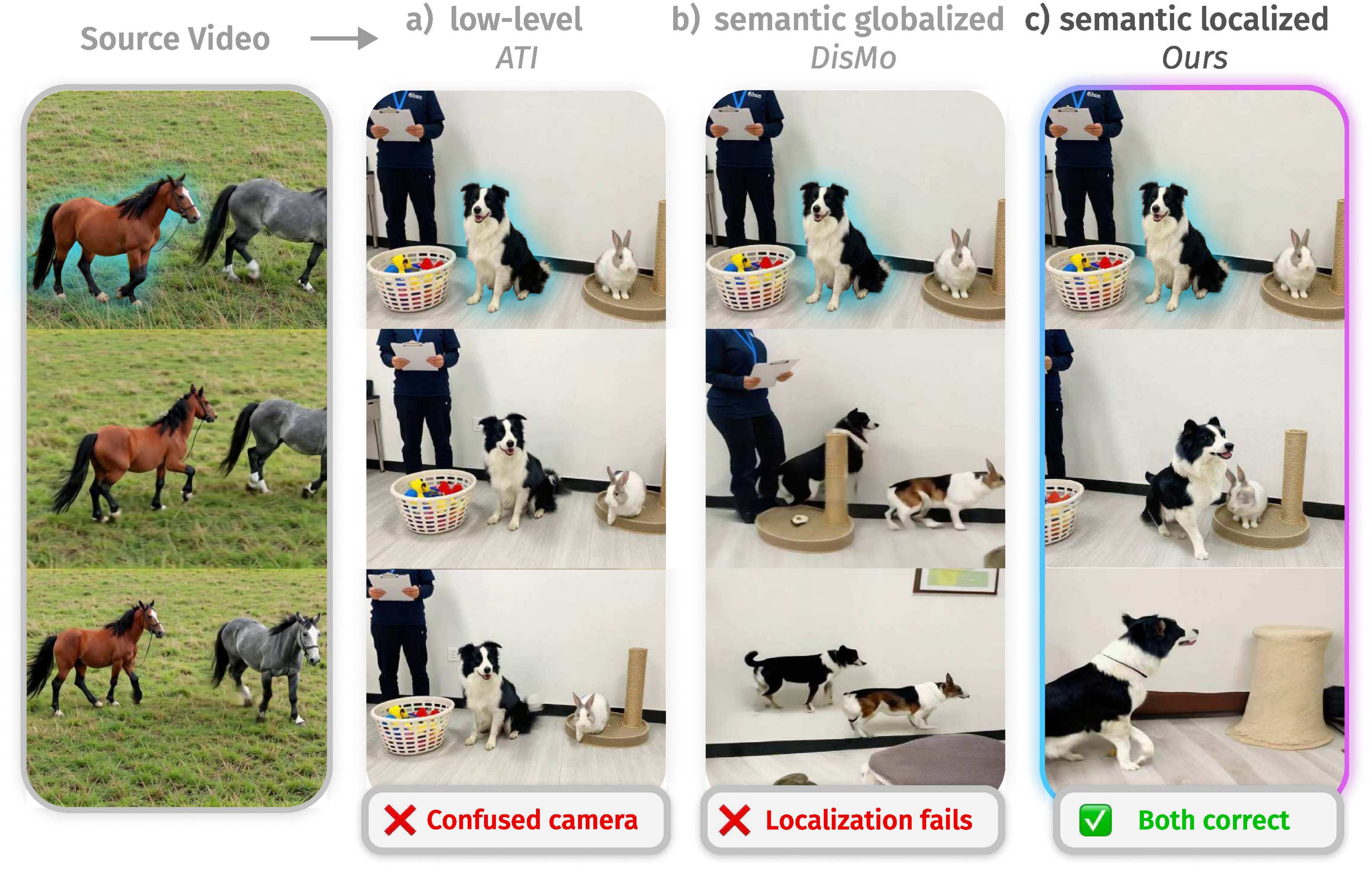}
    \caption{\textbf{Localized Motion Transfer.}
    Since the camera follows the subject, \textbf{(a)} confuses this for static subjects, \textbf{(b)} moves both subjects unrealistically, while \textbf{(c)} moves the correct subject with the right motion.
    }
    \label{fig:comp6}
\end{figure*}

Qualitative examples for the one-to-one motion transfer setting are shown in \cref{fig:comp1,fig:comp3,fig:comp4,fig:comp5,fig:comp6}. Prior methods exhibit several consistent failure modes in localized motion transfer. Low-level motion transfer approaches struggle under pose mismatch and often produce geometric inconsistencies and strong visual artifacts, such as clipping or abrupt scene changes. They also tend to entangle local object motion with global scene motion, causing subject motion to be mistaken for camera motion and vice versa. Global motion transfer methods instead fail to properly localize motion, frequently moving the wrong subject, affecting multiple instances, or hallucinating additional objects. Our method is substantially more robust to these issues, producing cleaner transfers that remain localized to the correct subject.


\subsection{Scene Composition}
\begin{figure*}[t]
    \centering
    \includegraphics[width=\linewidth]{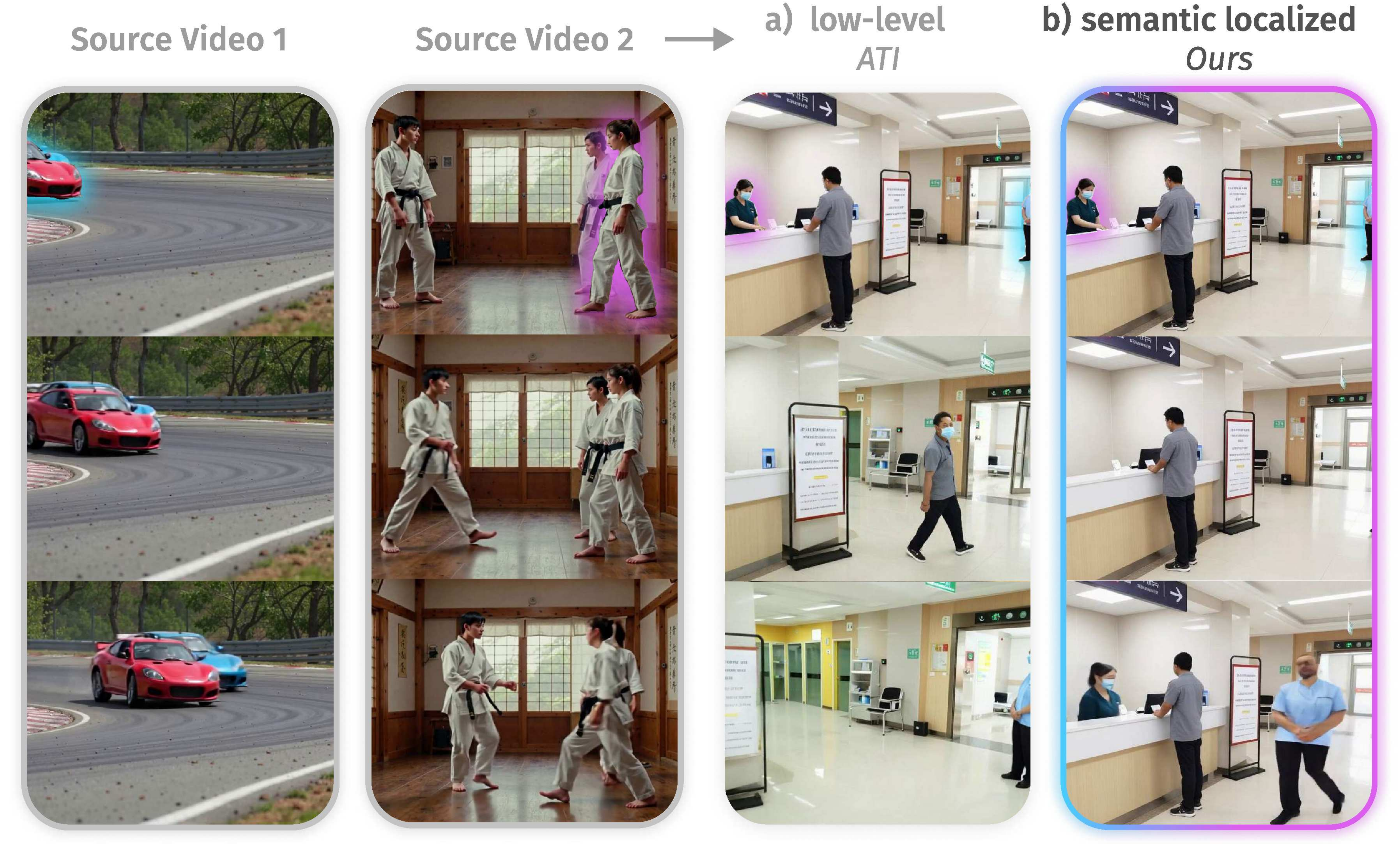}
    \caption{\textbf{Composite Motion Transfer.}
    \textbf{(a)} creates degenerate camera motion and does not transfer the correct motions \textbf{(b)} transfers the correct motion to the right subjects.
    }
    \label{fig:m2m_comp1}
\end{figure*}

\begin{figure*}[t]
    \centering
    \includegraphics[width=\linewidth]{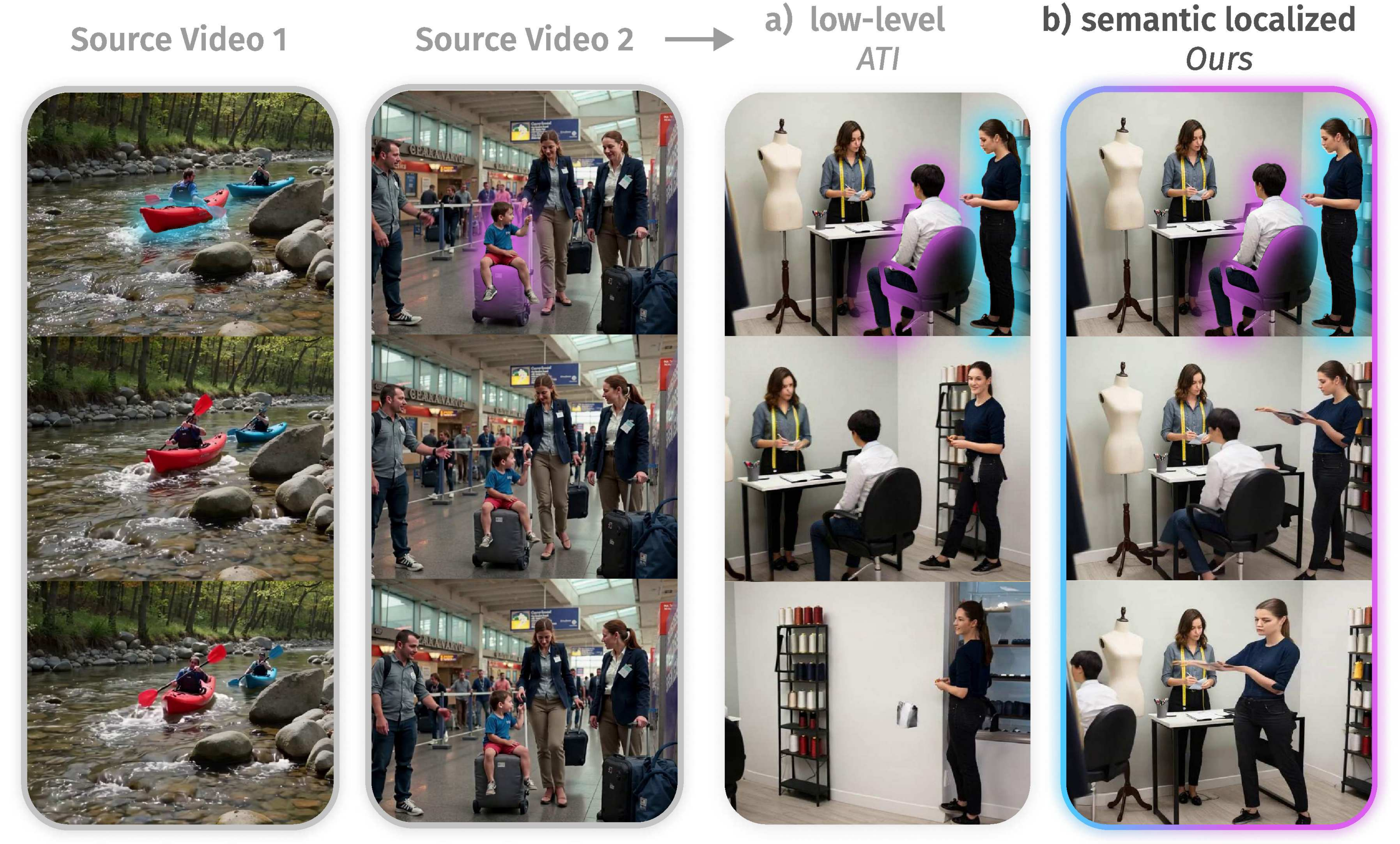}
    \caption{\textbf{Composite Motion Transfer.}
    \textbf{(a)} creates degenerate camera motion and does not transfer the correct motions, while \textbf{(b)} transfers the correct motion to the right subjects.
    }
    \label{fig:m2m_comp3}
\end{figure*}

\begin{figure*}[t]
    \centering
    \includegraphics[width=\linewidth]{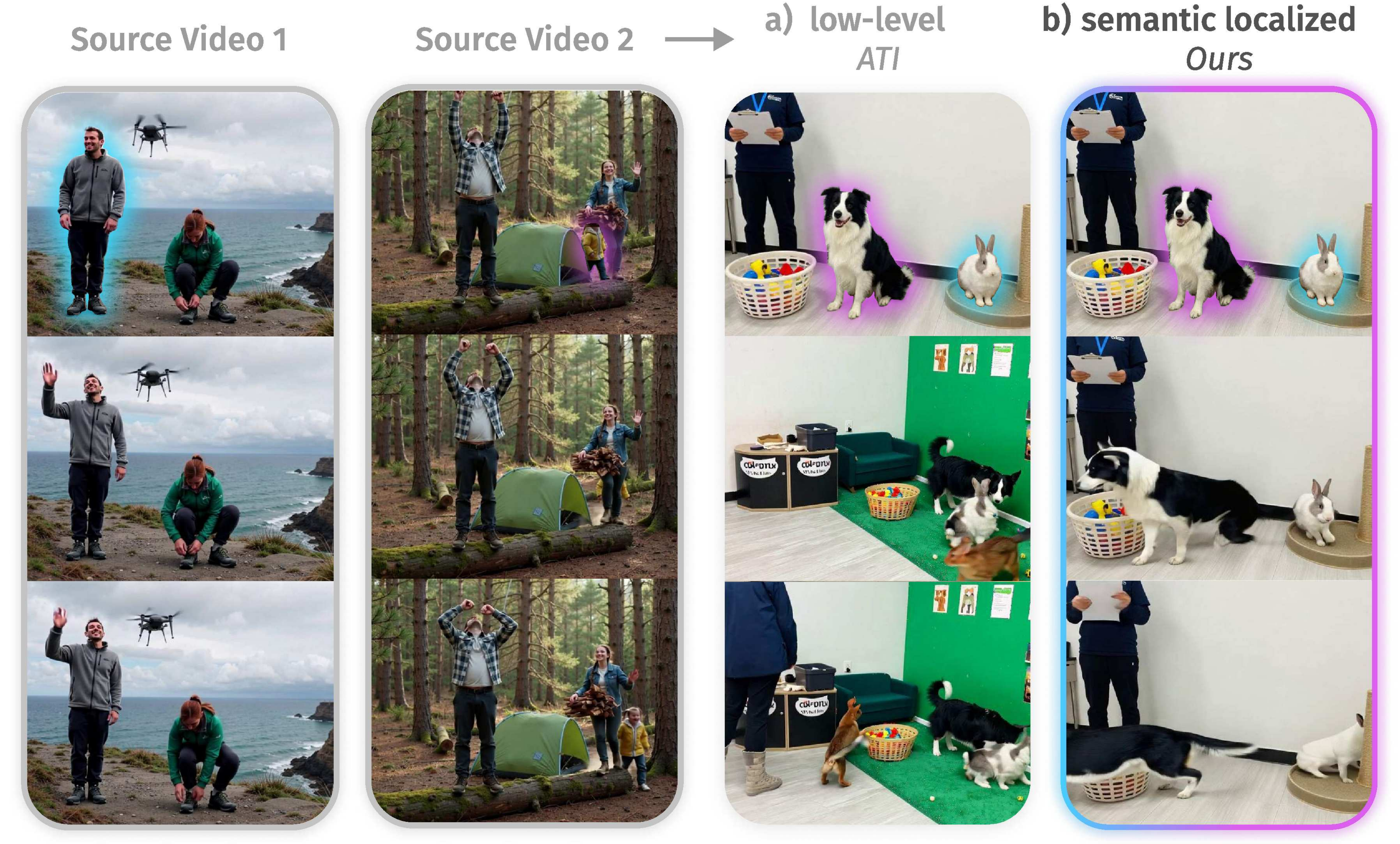}
    \caption{\textbf{Composite Motion Transfer.}
    \textbf{(a)} creates degenerate camera motion and does not transfer the correct motions, while \textbf{(b)} transfers the correct motion to the right subjects.
    }
    \label{fig:m2m_comp5}
\end{figure*}

\begin{figure*}[t]
    \centering
    \includegraphics[width=\linewidth]{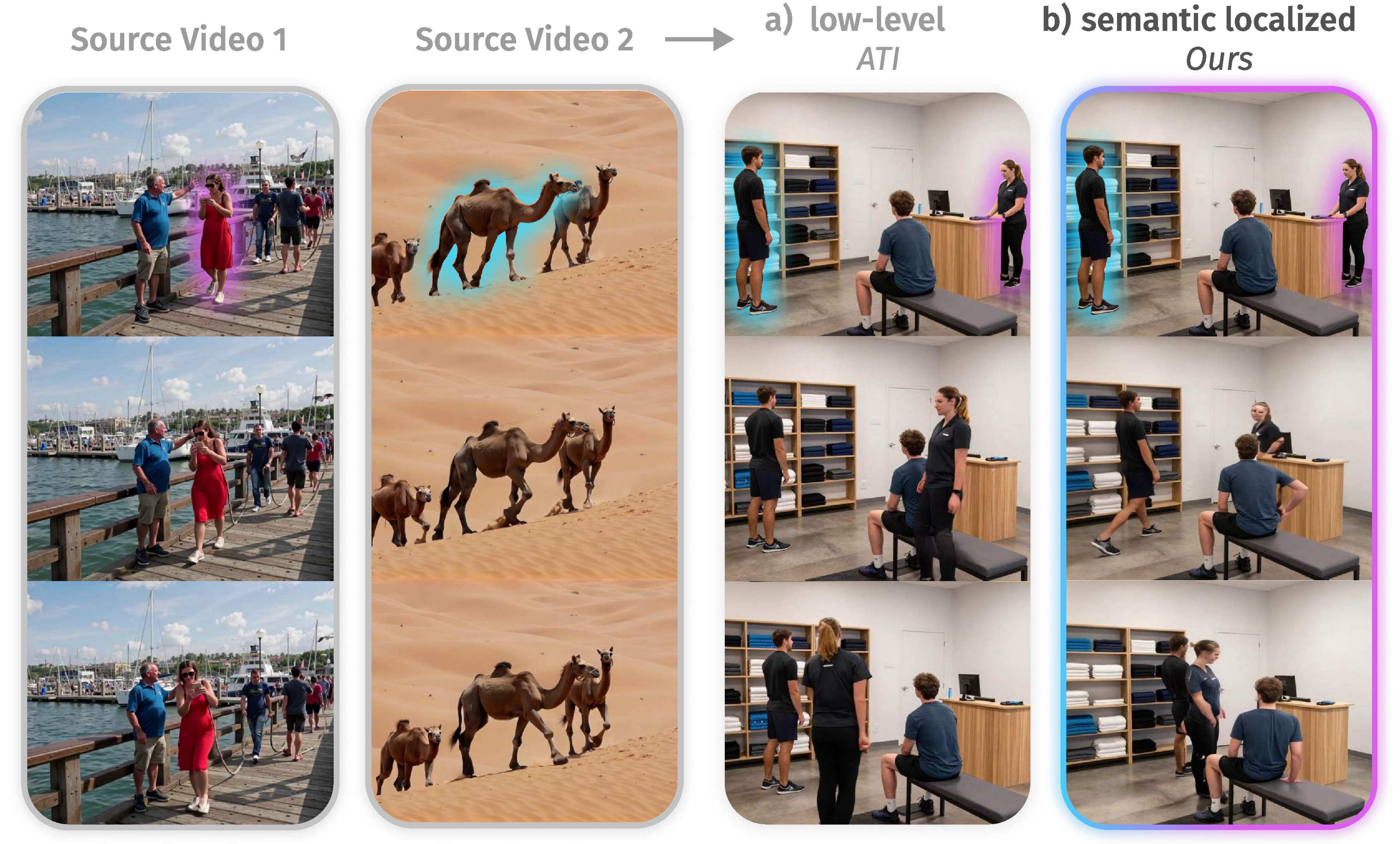}
    \caption{\textbf{Composite Motion Transfer.}
    \textbf{(a)} does not transfer the correct motions, while \textbf{(b)} transfers the correct motion to the right subjects.
    }
    \label{fig:m2m_comp6}
\end{figure*}

Qualitative examples for the compositional motion transfer setting, where multiple motion sources are mapped to multiple target objects, are shown in \cref{fig:m2m_comp1,fig:m2m_comp3,fig:m2m_comp5,fig:m2m_comp6}. We utilize low-level methods as a comparison for the composite motion transfer examples, global motion transfer models are not applicable, since their video generation pipeline does not accept multiple motion embeddings from different subjects.
In composite motion transfer, prior methods mainly fail by misassigning motions across subjects, omitting parts of the requested motion composition, or introducing degenerate camera motion. These errors indicate difficulty in jointly modeling multiple subject-specific motions while preserving a stable global scene. As a result, the generated videos often do not reflect the intended composition of motions. Our method better separates subject-level motion from global scene dynamics and therefore transfers the correct motions to the correct subjects more reliably.

\begin{figure*}[t]
    \centering
    \includegraphics[width=\linewidth]{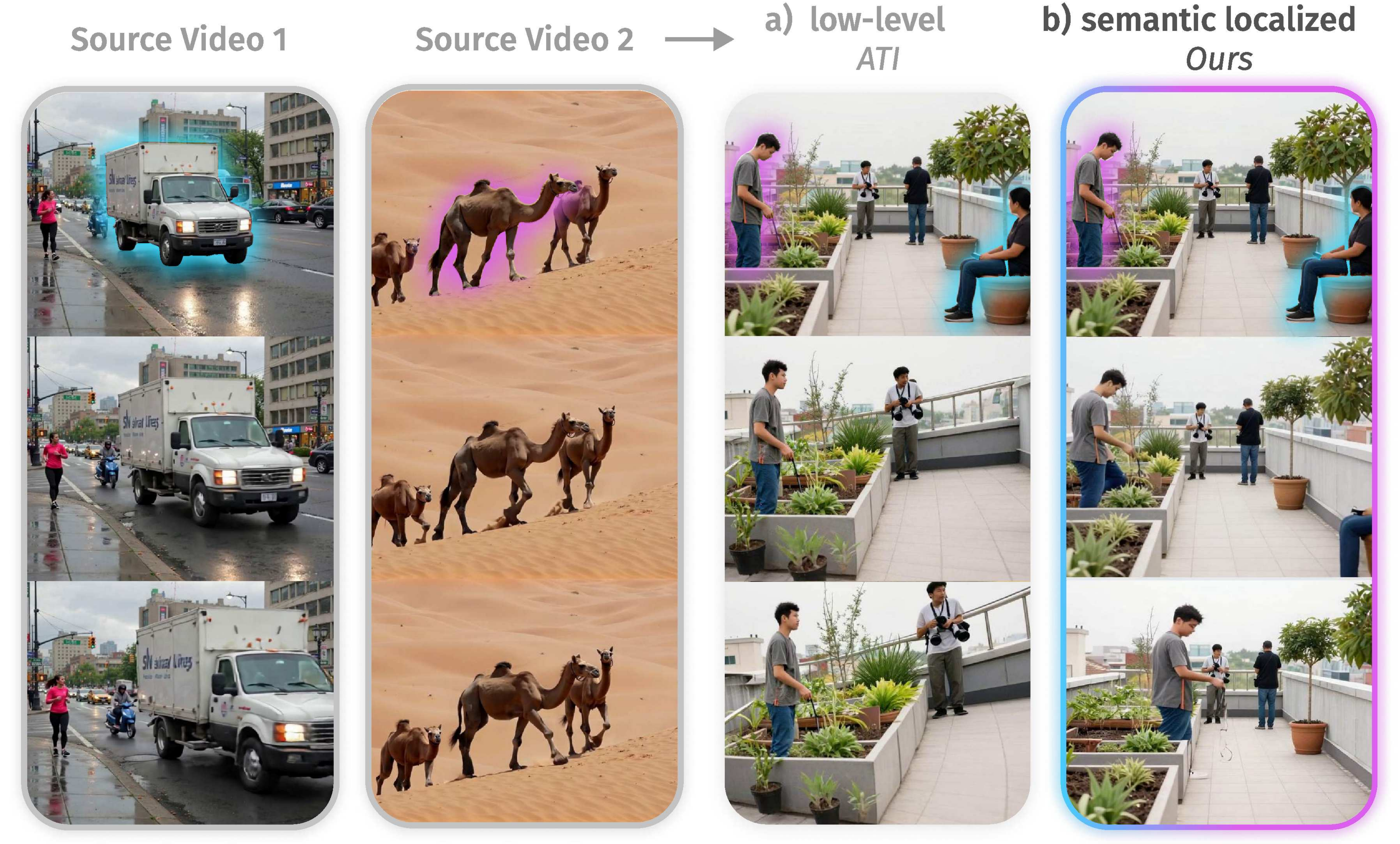}
    \caption{\textbf{Composite Motion Transfer.}
    \textbf{(a)} creates degenerate camera motion and does not transfer the correct motions, while \textbf{(b)} transfers the correct motion to the right subjects.
    }
    \label{fig:m2m_comp2}
\end{figure*}


\subsection{Text-prompted Generation}
\begin{figure}[b]
\centering
\includegraphics[width=\linewidth]{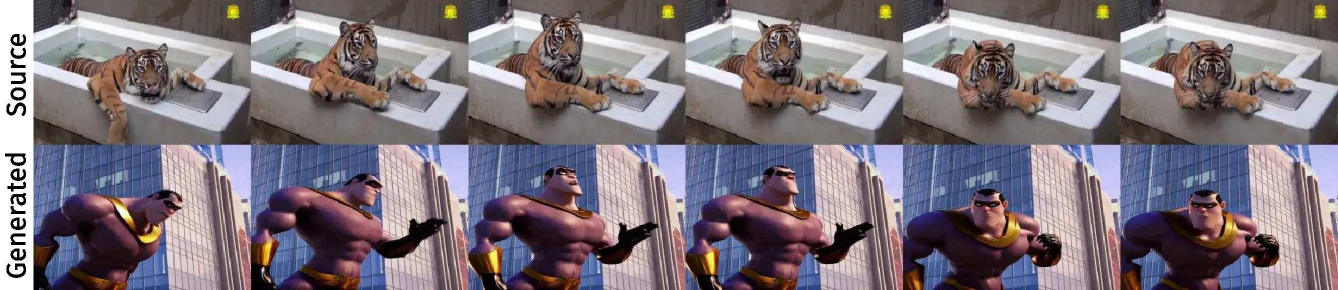}
\caption{\textbf{Text-prompted generation.} \textbf{Top:} Driving video providing the articulated motion representation. \textbf{Bottom:} Generated video conditioned on the extracted motion and a text prompt, without using a target start frame.}
\label{fig:rebuttal_promptable}
\end{figure}

While our primary experiments focus on motion transfer conditioned on a target start frame, the learned motion representation is not restricted to image-conditioned generation. We further demonstrate that it can support promptable text-only generation by fine-tuning the video model without start-frame conditioning. In this setting, the model receives a motion embedding extracted from a driving video together with a textual description of the desired scene. The text prompt determines the generated appearance and semantic content, while the motion embedding specifies the temporal dynamics of the generated video. As shown in \cref{fig:rebuttal_promptable}, the model transfers the articulated motion of the driving sequence to a novel subject and scene specified only through text. This demonstrates that the representation captures reusable motion information that can be combined with different appearance-conditioning modalities, extending its applicability beyond image-to-video motion transfer.

\subsection{User Study Details}
\label{supp:user_study}
We conducted a blinded user study on Prolific.com involving 18 participants and 17 non-cherrypicked test examples. For each example, participants were shown the generated videos without information about the underlying method or model identity. They independently evaluated each video along three dimensions: overall visual realism, the quality and spatial locality of the transferred motion, and the presence of visible artifacts. The transfer quality and locality criterion assessed both how faithfully the source motion was reproduced and whether it remained confined to the intended target subject without affecting unrelated regions of the scene.

\subsection{Ethical Impact}
As with most generative models, our approach is inherently dual-use. While it enables fine-grained control over object motion for creative, scientific, and educational applications, the same capabilities could be misused to fabricate deceptive or misleading video content. In particular, localized motion transfer may facilitate the manipulation of how individual subjects appear to move or act within a scene, potentially creating convincing visual evidence of events that never occurred. Responsible deployment should therefore include clear disclosure of generated content, provenance tracking where possible, and appropriate safeguards against deceptive use.